\documentclass[journal,10pt]{IEEEtran}

\usepackage[cmex10]{amsmath}
\usepackage{amssymb}
\usepackage{mathrsfs}
\usepackage{cite, color}
\usepackage{array}
\usepackage{enumerate}
\usepackage{graphicx}
\usepackage[usenames,dvipsnames]{xcolor}
\usepackage[caption=false,font=footnotesize]{subfig}

\newtheorem{theorem}{Theorem}
\newtheorem{lemma}{Lemma}

\newtheorem{remark}{Remark}

\newcommand{\R}{\mathbb{R}}

\newcommand{\E}{\mathbb{E}}

\newcommand{\calX}{\mathcal{X}}
\newcommand{\bcalX}{\mathbf{\calX}}

\newcommand{\calO}{\mathcal{O}}

\newcommand{\calK}{\mathscr{K}}

\newcommand{\calL}{\mathscr{L}}
\newcommand{\calD}{\mathscr{D}}

\newcommand{\kbar}{\bar{k}}

\newcommand{\Dhat}{\widehat{D}}
\newcommand{\calDhat}{\widehat{\calD}}

\newcommand{\calKhat}{\widehat{\calK}}
\newcommand{\Lhat}{\widehat{L}}
\newcommand{\calLhat}{\widehat{\calL}}

\newcommand{\Yhat}{\widehat{Y}}

\newcommand{\VAR}{\operatorname{Var}}
\newcommand{\COV}{\operatorname{Cov}}

\newcommand{\betadistrib}{\operatorname{Beta}}

\newcommand{\onevec}{\vec{\mathbf{1}}}

\begin{document}

\title{Laplacian Eigenmaps from Sparse, Noisy Similarity Measurements}

\author{Keith~Levin,~
        Vince~Lyzinski 
\thanks{Keith Levin is with the Department of Computer Science, Johns Hopkins University, Baltimore, MD, 21218 USA {\tt klevin@jhu.edu}}
\thanks{Vince Lyzinski is with the Department of Applied Mathematics and Statistics and the Human Language Technologies Center of Excellence, Johns Hopkins University, Baltimore, MD, 21218 USA {\tt vlyzins1@jhu.edu}}
\thanks{This work is partially supported by the
XDATA program of the Defense Advanced Research Projects Agency (DARPA)
administered through Air Force Research Laboratory contract FA8750-12-2-0303;
the DARPA SIMPLEX program SPAWAR contract N66001-15-C-4041;
and DARPA GRAPHS contract N66001-14-1-4028. 
}}

\maketitle

\begin{abstract}
Manifold learning and dimensionality reduction techniques are ubiquitous
in science and engineering, but can be computationally expensive procedures
when applied to large data sets or when similarities are expensive to compute.
To date, little work has been done to investigate the tradeoff between
computational resources and the quality of learned representations.
We present both theoretical and experimental explorations of this question.
In particular, we consider Laplacian eigenmaps embeddings based on
a kernel matrix, and explore how the embeddings behave when this kernel
matrix is corrupted by occlusion and noise.
Our main theoretical result shows that under modest noise and occlusion
assumptions, we can (with high probability)
recover a good approximation to the Laplacian eigenmaps embedding
based on the uncorrupted kernel matrix.
Our results also show how regularization can aid this approximation.
Experimentally, we explore the effects of noise and occlusion 
on Laplacian eigenmaps embeddings of two real-world data sets,
one from speech processing and one from neuroscience,
as well as a synthetic data set.
\end{abstract}


\section{Introduction and Motivation}
\label{sec:motivation}

\IEEEPARstart{M}{anifold-based} dimensionality reduction techniques
operate under the assumption that
data observed in a high-dimensional space
lie on a low-dimensional
manifold~\cite{TenSilLan2000,RowSau2000,BelNiy2003,BelNiySin2006}.
Owing to the ubiquity of large high-dimensional data sets,
these techniques have been well studied,
with applications across many disparate fields~\cite{van2009dimensionality}.
In addition to the classical linear techniques
(e.g., PCA~\cite{jolliffe2002principal}, MDS~\cite{CoxCox2001} and
CCA~\cite{hotelling1936relations, hardoon2004canonical}),
numerous manifold embedding procedures have been proposed to discover
intrinsic low-dimensional structure in nonlinear data
(e.g., ISOMAP~\cite{TenSilLan2000} and
Laplacian eigenmaps~\cite{BelNiy2003}, among others).
These nonlinear techniques typically
attempt to preserve some notion of local geometry in the embedding.  
As such, they tend to be empirically robust to modest noise and
outliers~\cite{BelNiy2003},
though general theoretical results in this direction are comparatively few.

Herein, we theoretically and practically explore the robustness
of Laplacian eigenmaps to very general noise conditions.
The present work differs from most manifold embedding robustness results
in two key ways:
first, we assume that the uncertainty lies not in the observations
themselves, but rather in our measurement of 
the pairwise similarities used to construct the kernel matrix.
Second, the noise model is entirely nonparametric:
we make no distributional assumptions on the noise other than
unbiasedness (see Equation~\eqref{eq:model} below).

\subsection{Problem Description}
Suppose that $\bcalX$ is a set of objects,
endowed with a notion of similarity captured by a kernel function
$\kappa : \bcalX \times \bcalX \rightarrow [0,1]$;
i.e., $x,y \in \bcalX$ are similar if $\kappa(x,y) \approx 1$,
and $x,y \in \bcalX$ are not similar if $\kappa(x,y) \approx 0$.
Given $n$ observations $x_1,x_2,\dots,x_n \in \bcalX$,
we can represent their similarities via a hollow
(i.e., no self-loops), undirected weighted graph
with adjacency matrix $\calK$ given by
\begin{equation} \label{eq:kernel}
\calK_{ij} = \begin{cases} \kappa(x_i,x_j) &\mbox{ if } i \neq j \\
				0 &\mbox{ otherwise. }\end{cases}
\end{equation}
Manifold-based dimensionality reduction techniques
seek to recover the
low-dimensional structure intrinsic in the similarities captured by $\calK$.
We note that some manifold embedding algorithms rely on
distance or disimilarity measures
rather than similarities, but the distinction is immaterial here.

The quality of the embedding of $\calK$
depends upon the quality of the similarity measure $\kappa$
and upon our ability to compute the similarity accurately.
If $\kappa$ only approximately captures the ``correct''
notion of similarity between observations,
it is natural to ask how this influences the quality of the embedding.
Similarly, when $\kappa(x,y)$ is expensive to compute,
we might ask whether an embedding of similar quality
is possible based on an inexpensive approximation
or by computing $\kappa(x,y)$ for only a fraction of
all pairs of observations, and inferring the rest of $\calK$,
for example, by applying Chatterjee's
universal singular value thresholding (USVT)~\cite{Chatterjee2015}.

The Laplacian eigenmaps embeddings constructed in~\cite{LevHenJanLiv2013}
serve as an illustrative example.
The authors' data consists of a set of $10,383$ word examples,
each represented by a time series of acoustic feature vectors.
For word examples $x_i$ and $x_j$,
the corresponding entry in the kernel matrix is
$$\calK_{ij} = \exp\{ -d^2(x_i,x_j)/\sigma^2 \},$$
where $d(x_i,x_j)$ is a function of
the dynamic time warping (DTW) alignment cost~\cite{SakChi1978}
between $x_i$ and $x_j$.
We refer the reader to~\cite{LevHenJanLiv2013} and references therein
for technical details.
The inadequacies of DTW as a word similarity measure are well documented in
the speech processing literature~\cite{LevHenJanLiv2013,LevJanVan2015}.
Additionally, DTW cost is computationally expensive, requiring time that
scales as the product of the lengths of the two aligned sequences.
As such, a fast estimate of
$d(x_i,x_j)$ or $\kappa(x_i,x_j)$ is acceptable,
and it is preferable to avoid computing all $O(n^2)$
alignments required to populate the kernel matrix.

\subsection{Our Model}
In light of the above, we consider the following model.
We assume a fixed set of observations $x_1,x_2,\dots,x_n \in \bcalX$,
and a similarity function $\kappa$ defined on $\bcalX \times \bcalX$,
giving rise to a true but unknown symmetric kernel matrix
$\calK=[\calK_{ij}]\in[0,1]^{n\times n}$.
The embedding learned from $\calK$
is the best embedding we could hope to learn,
in that it accurately and completely captures all the information available to
us about $x_1,x_2,\dots,x_n$.
The data processing inequality~\cite{CovTho2006} implies that
given the data, kernel function and  embedding procedure,
adding noise and occlusion to $\calK$ cannot improve
the embeddings from the standpoint of subsequent
inference or classification.
Suppose, however, that rather than observing $\calK$, we observe
a random symmetric matrix $Y \in \R^{n \times n}$, whose entries are
generated independently as
\begin{equation} \label{eq:model}
  Y_{ij} = Y_{ji} = \begin{cases} K_{ij} &\mbox{ with probability } p\\
                0 &\mbox{ with probability } (1-p), \end{cases}
\end{equation}
where the $K_{ij} \in [0,1]$ are independent random variables with
$\E K_{ij} = \calK_{ij}$
and $p \in [0,1]$ is the 
(expected) fraction of entries of $\calK$ that are observed.
We note that our results hold for similarity functions
bounded by any constant, and our use of the range $[0,1]$
is without loss of generality.
We can think of $K$ as a corrupted version of $\calK$, with errors
reflecting, for example, our failure to fully capture the correct notion of
similarity on $\bcalX$, or
approximation error arising from
estimating a computationally expensive $\kappa(x,y)$.
Similarly, we can view the sparsity of $Y$ as reflecting the fact that 
when $n$ is large or $\kappa$ is expensive to compute,
we would like to
avoid computing all $O(n^2)$ pairwise similarities in $\calK$.
Our model is meant to account for general uncertainty in the
kernel matrix, which may come from many sources
(e.g., computational restrictions, estimation, etc.).
Ultimately, we require only that errors be entry-wise independent
and unbiased.

When $\calK_{ij} \approx 0$ or $\calK_{ij} \approx 1$,
our model allows $K_{ij}$ very little variance.
In many applications, the cases when $\kappa(x,y) \approx 0$ or
$\kappa(x,y) \approx 1$ are less prone to error,
which is reflected in our model.
Indeed, it is often easy to detect when two
observations are very similar or very dissimilar, whereas one expects higher
variance in estimation of similarity when, say, $\kappa(x,y) = 1/2$.

\begin{remark}[Error Generalization]
Our model is a good approximation to more complicated error models.
As an example, consider the Gaussian kernel
$\kappa(x,y) = \exp\{ -d^2(x,y)/\sigma^2 \}$,
where $\sigma > 0$ is the kernel bandwidth.
A more natural but less tractable error model is one in which
$D_{ij}$ is an estimate (possibly biased) of $d(x_i,x_j)$
and our kernel matrix is $K_{ij} = \exp\{ -D^2_{ij}/\sigma^2 \}$,
say, $D_{ij} = d_0 + E_{ij}$ where $E_{ij}$ is a random error term.
A Taylor expansion of $\exp\{ -t^2/\sigma^2 \}$ about $d_0 = d(x_i,x_j)$
shows that
(taking $\sigma=1$ without loss of generality
and using the fact that $\calK_{ij} = e^{d_0^2}$)
$$ K_{ij} = \calK_{ij}
	- 2d_0 e^{-d_0^2} E_{ij}
       + (4d_0^2 - 2)e^{-d_0^2} E_{ij}^2
	+ O( E_{ij}^3 ). $$
We see that so long as $E_{ij}$ is reasonably well-behaved, we still have
$\E K_{ij} \approx \calK_{ij}$, and an approximate version of the results
presented in this paper will hold.
More broadly, we note that
so long as $|\E K_{ij} - \calK_{ij}|$ is suitably small
for most entries, our results can be extended to the case
of biased errors.
These observations are borne out by experiment
(See Figures~\ref{fig:HDSmult} and~\ref{fig:HDSbias}).
\end{remark}

In this paper, we theoretically and practically
explore under what conditions it is suitable to use
the embedding learned using $Y$ in place of $\calK$.
Under such conditions,
we can obtain embeddings with quality comparable
to those produced from $\calK$, at a greatly reduced computational cost.
In the present work, we consider the performance of
Laplacian eigenmaps~\cite{BelNiy2003,BelNiySin2006} under this model, 
though we believe that the results extend to other
embedding techniques, as well.

\subsection{Laplacian Eigenmaps}
As originally described in~\cite{BelNiy2003,BelNiySin2006},
Laplacian eigenmaps embeds the observed data $\bcalX$
into $\R^d$ by first constructing the $k$-nearest-neighbor ($k$-NN) or
$\epsilon$-graph $G=(V,E)$ from $\bcalX$.
In the $k$-NN graph, an edge is present between $i$ and $j$
if $x_i$ is among the $k$ nearest neighbors (according to some distance
defined on $\bcalX$)
of $x_j$ or vice versa.
In the $\epsilon$-graph, $i$ and $j$ are adjacent if $\|x_i-x_j\|^2<\epsilon$
for a given threshold parameter $\epsilon$.
We define $W$, the weighted adjacency matrix of $G$,
by $$W_{ij}=\begin{cases} \calK_{ij} &\mbox{ if } \{i,j\}\in E\\
                0 &\mbox{ else}, \end{cases}$$
and let $\calD\in\mathbb{R}^{n\times n}$ be the diagonal matrix defined by
$\calD_{ii}=\sum_{j}W_{ij}$ for $i\in[n]$.
Then the normalized weighted graph Laplacian of $G$~\cite{Chung1997}
is given by $\calL(W)=\calD^{-1/2}W\calD^{-1/2}$.
If the eigendecomposition of $\calL(W)$ is given by 
$\calL(W)=U\Lambda U^\top$ with the diagonal entries of
$\Lambda$ nonincreasing,
then Laplacian eigenmaps embeds $\bcalX$ via
$U[:,2:d+1]$---the first $d$ nontrivial eigenvectors of $\calL(W)$.
(note that $U[:,1]=\onevec$, the trivial all-ones vector).
This embedding optimally
preserves the local geometry of $\bcalX$ in a least squares sense.

In the event that $\calK$ is noisily
and incompletely observed as $Y$,
how does the $d$-dimensional Laplacian eigenmaps embedding of $Y$
compare with that of $\calK$?
Our main result, Theorem~\ref{thm:main},
deals with the regularized matrix $[Y_{ij} + r]$ rather than
$Y$ itself, owing to the fact that when $p$ is small,
the matrix $p\calK = \E Y$ may be quite sparse,
in the sense that some or all of the row sums $\sum_{j=1}^n p\calK_{ij}$
are too small to guarantee necessary concentration
inequalities~\cite{Oliviera2010,Tropp2012,LeLevVer2015}.
Regularization prevents this pitfall, at the cost of changing the
matrix to which we converge.
We discuss regularization at more length in
Subsection~\ref{subsec:mxconcentration}.
Intuitively, our main theorem states that
the embedding produced from a regularized version of $Y$ is
similar to that produced by $\calK$.
This implies that we can avoid the $O(n^2)$ exact computations for $\calK$,
using instead the potentially less computationally expensive $Y$,
with little loss in downstream performance.
\begin{remark}
We depart from Laplacian eigenmaps as originally described~\cite{BelNiy2003}
in that we do not build a
$k$-NN graph or $\epsilon$-graph from $\bcalX$.
However, a suitably-chosen kernel function (e.g., the Gaussian
kernel) ensures that $\calK$ approximates a $k$-NN or $\epsilon$-graph,
with $Y$ a noisily-observed subgraph of $\calK$.
\end{remark}

\subsection{Notation}
For a set $S$, we denote the complement of $S$ by $S^c$.
For a matrix $B \in \R^{n \times n}$, we let
$\lambda(B)$ denote the multi-set of eigenvalues of $B$,
and for $S \subset \R$, we define $\lambda_S(B) = \lambda(B) \cap S$.
We let $J \in \R^n$ denote the matrix of all ones.

We make use of standard big-$O$ notation, writing $f(n) = O(g(n))$
to mean that there exists a constant $C > 0$ such that
$f(n) \le Cg(n)$ for suitably large $n$.
Similarly, we write $f(n) = o(g(n))$ to mean that
$f(n)/g(n) \rightarrow 0$ as $n \rightarrow \infty$.
We use $f(n) = \Omega(g(n))$ to denote that
$f$ grows at least as quickly as $g$ does,
i.e., to denote that $g(n) = O(f(n))$,
and we write $f(n) = \omega(g(n))$ when $g(n) = o(f(n))$.

Throughout this paper,
all quantities are assumed to depend on $n$,
a fact that we highlight by subscripting or superscripting
with $n$ (e.g., $\calK = \calK^{(n)}$),
but which we will suppress in much of the text for ease of notation.
Our main theorem, Theorem~\ref{thm:main}, is a finite-sample result,
with $\calK^{(n)}$ viewed as fixed for each $n$,
and $K^{(n)}$ and $Y^{(n)}$ randomly generated from $\calK^{(n)}$.
We note that all of our results can be restated as holding almost surely
as $n \rightarrow \infty$
by assuming suitable lower bounds on the constants
in the supporting Lemmas
so as to ensure that the probabilities of the various
``bad events'' are summably small.
An application of the Borel-Cantelli lemma then
implies that our desired events hold almost surely.
This modification can be made to work either in the case
(a) where we view $Y, K$ and $\calK$ as (growing, ``nested'')
principle submatrices of infinite matrices,
or (b) in the case where we consider a sequence of
fixed matrices $(\calK^{(n)})_{n=1}^\infty$.

In this work, we assume $\calK$ to be fixed
(i.e., not random-- the randomness lies entirely in $Y$ and $K$).
This assumption is made primarily for the sake of brevity and simplicity,
since randomness in $\calK$ would have to come from random selection
of the sample $x_1,x_2,\dots,x_n \in \bcalX$ according to some distribution
$F$ on $\bcalX$.
Clearly, the properties of $\calK$ depend on the properties of
$F$ and $\bcalX$, but a thorough exploration of precisely how $F$ and $\bcalX$
influence $\calK$ is beyond the scope of this paper,
and we leave it for future work.

\subsection{Roadmap}
Section~\ref{sec:relatedwork} surveys
robust manifold-based dimensionality reduction and related problems.
We present our theoretical results in Section~\ref{sec:mainresults},
and explore these results experimentally in Section~\ref{sec:experiments}.
We close with a brief discussion in Section~\ref{sec:discussion}.

\section{Related Work}
\label{sec:relatedwork}

\subsection{Manifold Learning}
Manifold learning is a general class of techniques for
nonlinear dimensionality reduction that seek to
embed a collection of observations into Euclidean space in a way that
preserves some aspect of the structure of those observations.
For example, given a collection of objects and some notion of distance on
those objects, we may wish to embed the objects into Euclidean space in such
a way that all pairwise distances are (approximately)
preserved~\cite{Indyk2001,Linial2002}.
A host of different embedding techniques have been proposed in the literature
(see, for example,~\cite{RowSau2000,TenSilLan2000,CoxCox2001,HinRow2002,DonGri2003,CoiLaf2006})
to preserve the numerous different notions of structure in the data.
As outlined in~\cite{YanXuZhaZhaYanLin2007},
it is possible to view many of these approaches as special cases of
a more general framework

There is a large amount of literature dedicated to improving the performance of
manifold learning and dimensionality reduction algorithms
in the presence of noise and missing data; see,
for example,~\cite{ChaYeu2006,HeiMai2006,CanLiMaWri2011,ShaKalBreBroVan2015}.
The present work differs from most such results in
the following key ways:
We assume that the uncertainty lies not in the observations
themselves, but rather in the computation of 
the pairwise similarities or distances used to construct the kernel matrix,
and our model of this uncertainty is nonparametric.
Additionally, we make no assumption that the
observations lie in Euclidean space.
Rather, the objects under study are arbitrary
(e.g., they may be time series, graphs, etc.),
and information about the geometry of $\bcalX$
comes through the kernel function $\kappa$.

With the rise of big data and the continued
popularity of kernel methods, much research has gone toward
faster construction and embedding of the kernel matrix by speeding up the
evaluation of the kernel function itself~\cite{WilSee2001,LeSarSmo2013},
the embedding procedure~\cite{baglama2005augmented,brand2006fast},
and construction of the kernel matrix as a whole~\cite{FinSch2001}.
Construction of the kernel matrix is often
the major bottleneck in machine learning
systems~\cite{HofSchSmo2008,LevHenJanLiv2013,LevJanVan2015}.
In our model, embedding the partially observed noisy kernel matrix $Y$
allows for potentially dramatic speedups
compared to the computation of the full, clean kernel $\calK$.
A similarly-motivated idea was explored in~\cite{CheFanSaa2009},
where the authors
presented a pair of divide-and-conquer algorithms for approximately
constructing $k$-NN graphs on observations in Euclidean space. 
However, unlike our approach,
they do not consider noise in the observations themselves
or in the assessment of distances between observations.

Another close analogue to our present work is~\cite{RohChaYu2011},
in which the authors
theoretically and empirically explored the robustness properties of
spectral clustering: i.e., Laplacian eigenmaps applied to a binary
adjacency matrix followed by $k$-means clustering.
In the language of the present paper,
they considered the inner product kernel matrix
$\calK\in\mathbb{R}^{n\times n}$
on a fixed (but unknown) subset $\bcalX\subset\mathbb{R}^d$.  
From this kernel, they observed the matrix
$Y\in\{0,1\}^{n\times n}$ with independent entries
\begin{equation} \label{eq:modelRCY}
  Y_{ij} = Y_{ji} = \begin{cases} 1 &\mbox{ with probability } \calK_{ij}\\
                0 &\mbox{ with probability } (1-\calK_{ij}). \end{cases}
\end{equation}
They 
compared the Laplacian spectral embedding based on $\calK$
with that based on $Y$.
Their key result showed that, under some mild assumptions
on the spectrum of $\calL(\calK)$ (the normalized Laplacian of $\calK$),
the eigenspace of $\calL(Y)$ does not significantly differ from the
corresponding eigenspace of $\calL(\calK)$ (after suitable rotation).
As a result, they prove that spectral clustering of $\calL(Y)$
consistently estimates the clusters obtained
by spectrally clustering $\calL(\calK)$.
While our main theorem uses results
(\!\!\!\cite[Prop. 2.1 and Thm. 2.2]{RohChaYu2011}) developed in that paper,
the generality of our occlusion model~\eqref{eq:model}
compared to~\eqref{eq:modelRCY} requires new proof techniques.
Additionally, our manifolds do not necessarily have a
well-defined cluster structure
(as the stochastic blockmodel graphs of \cite{RohChaYu2011} do),
and so we do not consider consistency of clustering of our embedding.
Rather, in Theorem~\ref{thm:main},
we prove that the relevant eigenvectors of $\calL(Y)$ do not significantly
differ from the corresponding eigenvectors of $\calL(\calK)$.  
As in~\cite{RohChaYu2011}, we expect the consistency of subsequent inference
to similarly follow.

\subsection{Matrix Completion and Data Imputation}
A natural approach to applying Laplacian eigenmaps to $Y$
is to first impute the missing entries of $Y$
using matrix completion techniques.
For example, with the additional assumption that $\calK$
is approximately low-rank,
it would be possible to impute the missing data via the
techniques developed in compressed sensing~\cite{CanRec2009}.
While some compressed sensing papers
have considered matrix completion in the presence of
both noise and occlusion~\cite{CanPla2009,CheJalSanCar2013},
most also require bounds on the incoherence of matrix $\calK$, 
a requirement that need not hold in general for the kernel matrices
we consider here.

Some matrix completion work has considered
imputing missing entries
in a distance matrix~\cite{AlfKhaWol1999,Trosset2000,JavMon2013}.
Among these works,~\cite{JavMon2013}
is closest in spirit to the problem considered in the present work.
In~\cite{JavMon2013}, the authors considered the problem of
placing $n$ objects into $d$-dimensional Euclidean space based on
noisy, occluded measurements of the $O(n^2)$ pairwise distances.
Their semidefinite programming-based approach
solves this problem under a very general error model,
where nothing is known about the errors other than a bound on their magnitude.
However, their model differs from ours in two key ways.
First, the observations in question are assumed to lie
in $d$-dimensional Euclidean space,
while ours need only be endowed with a kernel function.
Second, they assume that distance measurements are taken on all
pairs of points within a fixed radius of one another.
However, under our model, all entries of $\calK$ are equally
likely to be (noisily) observed.

Chatterjee~\cite{Chatterjee2015} considered the problem of completing
an arbitrary matrix based on partial, noisy observations,
with no specific assumptions on the matrix structure.
His universal singular value thresholding (USVT) procedure
constructs a minimax optimal estimate for $\calK$
based on its occluded, noisy measurement $Y$ (as defined in~\eqref{eq:model}).
Though we believe that the results obtained in this paper would hold
in a qualitatively similar way if we used USVT applied to matrix $Y$
prior to embedding, analyzing the behavior of the USVT estimate of
$\calK$ under the graph Laplacian is 
theoretically challenging, and we do not pursue it further here.
In empirical comparisons, we found our method and Chatterjee's USVT
performed nearly identically across our data sets.
We do note that USVT requires an expensive SVD computation,
and yields a dense matrix as an estimate of $\calK$, instead of
the sparse $Y$, which may be computationally intractable for large $n$.

\subsection{Matrix Concentration}
\label{subsec:mxconcentration}
Recent years have seen a flurry of results proving concentration
results for sums of random
matrices~\cite{Oliviera2010,Tropp2012,ChaChuTsi2012,AmiChaBicLev2013,JosYu2013,QinRoh2013,LeLevVer2015,LeVer2015},
in the spirit of the well-established scalar
analogues~\cite{ChuLu2006}.
Many existing concentration results require assumptions
about the density of the underlying graphs~\cite{RohChaYu2011,Oliviera2010}.
For example, many such results hold only in the dense regime and require a
lower bound on the average degree
(i.e., a lower bound on the row sums of the expected
value of the random matrix).
It is well known that the high variance associated with small average degree
precludes concentration of the Laplacian for general
weighted graphs~\cite{ChuLuVu2003,FeiOfe2005,LeLevVer2015,KloTsyVer2015}.
This is an issue for the problem considered in the present work,
especially when we observe only a small fraction of the matrix entries.

Existing empirical and theoretical results show that regularization
yields the desired concentration of the graph Laplacian for sparse graphs
(see~\cite{ChaChuTsi2012,AmiChaBicLev2013,JosYu2013,QinRoh2013,LeLevVer2015,LeVer2015} and references therein).
This regularization typically takes the form of either adding a small
number to each entry of the adjacency matrix, as in~\cite{LeLevVer2015},
or by adding to the degree matrix directly, as in~\cite{QinRoh2013}.
Our result draws on this line of work by investigating the behavior of the
Laplacian eigenmaps embeddings when regularization is applied.
In this sense, the current work is a natural outgrowth of~\cite{RohChaYu2011}
and~\cite{LeLevVer2015} in that the former
considers concentration of the Laplacian eigenmaps embeddings under
the Frobenius norm, and the latter considers concentration of
the regularized graph Laplacian under the spectral norm.
We follow the former of these two works and consider concentration under the
Frobenius norm, rather than spectral norm.
This differs from the bounds established
in~\cite{Oliviera2010,Tropp2012,LeLevVer2015,LeVer2015},
which show concentration of the adjacency matrix and graph Laplacian
under the spectral norm.
We prefer the Frobenius norm formulation of Theorem~\ref{thm:main},
as the Frobenius norm between the (suitably rotated)
eigenspaces has a natural interpretation as the Procrustes alignment
error of the orthogonal bases of the two different embeddings.

\section{Main Results}
\label{sec:mainresults}
Our goal is to theoretically and empirically understand the
impact of observation error on the embedding obtained via Laplacian eigenmaps.
That is, how much does the embedding obtained using matrix $Y$
degrade with respect to that obtained using matrix $\calK$?
We prove that
Laplacian eigenmaps is indeed robust to certain amounts of both occlusion and
noise by first proving that (a suitably regularized version of)
$\calL^2(Y)$ concentrates about (a regularized version of) $\calL^2(p\calK)$,
where $Y$ and $p$ are defined as in Equation~\eqref{eq:model}.
Combining this result with the
Davis-Kahan theorem~\cite{DavKah1970}, we obtain
in Theorem~\ref{thm:main} a guarantee that
the embedding learned from the occluded noisy kernel matrix is
similar (up to rotation) to that learned from
the regularized clean kernel matrix.
We provide relevant details below and in the appendix.

Let $G = (V,E)$ be an undirected,
loop-free, weighted graph on $n$ vertices
with edge weights $w_{ij} \ge 0$.
We represent $G$ by its adjacency matrix $A \in \R^{n \times n}$,
with entries
\begin{equation*}
A_{ij}=A_{ji}=\begin{cases}w_{ij}&\text{ if }\{i,j\}\in E\\
0&\text{ if }\{i,j\}\notin E.
\end{cases}
\end{equation*}
Given $A$, we define its normalized graph Laplacian by
$$ \calL(A) = \calD(A)^{-1/2} A \calD(A)^{-1/2}, $$
where $\calD(A) \in \R^{n \times n}$, the degree matrix,
is diagonal with $\calD(A)_{ii} = \sum_{j=1}^n A_{ij}$
and inverse square root defined as
$$ \left(\calD(A)^{-1/2} \right)_{ii} = \begin{cases}
			1/\sqrt{ \calD(A)_{ii} } &\mbox{ if }
				\calD(A)_{ii} \neq 0 \\
			0 &\mbox{ otherwise.}
			\end{cases} $$
We note that the graph Laplacian as we have defined it differs from the
more commonly used $I - \calD(A)^{-1/2} A \calD(A)^{-1/2}$
(e.g., in~\cite{Chung1997}).
We will be interested in the eigenspace of $\calL(A)$,
and one can easily check that both our $\calL(A)$
and the more commonly used definition have the same eigenspaces.

In general, neither the adjacency matrix nor the graph Laplacian of
sparse random graphs concentrate about their means
owing to high variance in degree
distributions~\cite{FeiOfe2005,LeLevVer2015, LeVer2015}.
This suggests that we should not expect that $\calL(Y)$ will concentrate
for arbitrary kernel matrices, and hence we turn to regularization.
Let $J \in \R^{n \times n}$ denote the matrix of all ones.
Our main result will require us to bound
$\| \calL^2(Y + rJ) - \calL^2(p\calK + rJ) \|_F,$
where $Y$ is the sparse, noisy version of $\calK$ as specified
in~\eqref{eq:model},
and $r \ge 0$ is a regularization parameter.
We deal with the squared Laplacians for reasons discussed
in~\cite[Section~2]{RohChaYu2011}.
Namely, we require that
$\calL(Y+rJ)$ converge to $\calL(p\calK + rJ)$ in Frobenius norm.
To ensure convergence for a suitably broad class of matrices,
we must instead consider the squared Laplacians in combination with
the following Lemma, proved in~\cite{RohChaYu2011},
which ensures that if certain eigenvectors
of $\calL^2(Y+rJ)$ converge, then so do the relevant
eigenvectors of $\calL(Y+rJ)$.
\begin{lemma}[\!\!{\cite[Lemma 2.1]{RohChaYu2011}}]
Let $B \in \R^{n \times n}$ be symmetric.
\begin{enumerate}
\item $\lambda^2$ is an eigenvalue of $B^2$ if and only if
	either $\lambda$ or $-\lambda$ is an eigenvalue of $B$.
\item If $Bx = \lambda x$, then $B^2 x = \lambda^2 x$.
\item If $B^2 x = \lambda^2 x$, then
	$x$ can be written as a linear combination of eigenvectors of
	$B$ with corresponding eigenvalues $\lambda$ or $-\lambda$.
\end{enumerate}
\end{lemma}

Our main theorem, Theorem~\ref{thm:main}, shows that the
span of the eigenvectors corresponding to the largest eigenvalues
of the
Laplacian of $\calK$
and the Laplacian of the sparse noisy kernel matrix $Y$ are close.
As a consequence, subsequent inference performed on the
Laplacian eigenmaps embeddings
will be robust to the errors introduced in $Y$, since the
embeddings will be (nearly) isometric to one another.
In the statement of the theorem, we include subscript or superscript $n$ on all
quantities that depend on $n$, though we will drop these subscripts in the
sequel for notational convenience.
Recall that for $B \in \R^{n \times n}$, $\lambda(B)$ denotes the
 multi-set of eigenvalues of $B$ and for $S \subset \R$,
we define $\lambda_S(B) = \lambda(B) \cap S.$

\begin{theorem}  \label{thm:main}
Under the model described in~\eqref{eq:model},
for an open interval $S_n \subset \R$,
let $k_n = |\lambda_{S_n}(\calL(Y^{(n)} + r_n J))|$
be the cardinality of $\lambda_{S_n}(\calL(Y^{(n)} + r_n J))$
(counting multiplicities), and let $X_n \in \R^{n \times k_n}$
be the matrix whose columns form an orthonormal basis for the subspace
spanned by the eigenvectors of $\calL(Y^{(n)} + r_n J)$
with corresponding eigenvalues in
$\lambda_{S_n}(\calL(Y^{(n)} + r_n J))$.
Let $\kbar_n = |\lambda_{S_n}(\calL(p\calK^{(n)} + r_n J))|$
and let $\calX_n$ be the analogue of $X_n$
for $\calL(p\calK^{(n)} + r_n J)$.
Define 
\begin{equation} \label{eq:deltadef}
 \delta_n = \inf\{ |\ell - s|:
		\ell \in \lambda_{S_n^c}(\calL(p\calK + r_n J)), s \in S_n \} .
\end{equation}

Let $r_n$ depend on $n$ in such a way that
$r_n \ge n^{-1} \log n$ for suitably large $n$.
There exist constants $C,c>0$
and a positive integer $N$ such that $n \ge N$
implies that $k_n = \kbar_n$,
and there exists orthonormal rotation matrix $\calO_n$ such that
with probability at least $1-n^{-c}$,
$$ \| X_n - \calX_n \calO_n \|_F
\le C \left( \frac{ \log^{1/2} n }{ \delta_n r_n n^{1/2} } \right). $$
\end{theorem}
\begin{IEEEproof}
Combining Theorems~\ref{thm:RCYDK} and~\ref{thm:Lbound}
yields the result.
\end{IEEEproof}

\begin{figure}[ht]
  \centering
    \includegraphics[width=0.6\columnwidth]{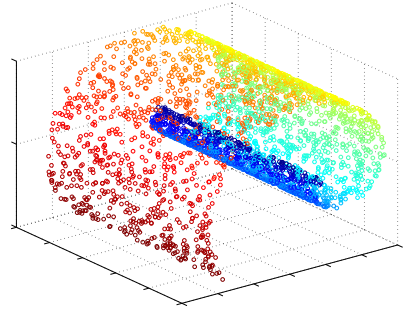}
  \caption{ Points sampled from a 3-dimensional swiss roll. }
  \label{fig:3Dswiss}
\end{figure}

\begin{remark} 
A key difference between the main theorem in~\cite{RohChaYu2011}
and our result is that we do not require a restriction on
the degrees of $p\calK$ directly.
Rather, we use regularization to ensure that no row sum is too small.
We note that letting $p=1$ and making minor adjustments to the arguments
in our concentration inequalities (namely, lower bounds on the entries of
the degree matrix $\calD$), we recover the main result
of~\cite{RohChaYu2011}, with a slightly better convergence rate.
Namely, if we define $\tau = n^{-1} \min_{i \in [n]} \calD_{ii}$,
our result has $\tau^{-1}$
controlling to rate of convergence of the eigenspaces
rather than $\tau^{-2}$ as in~\cite{RohChaYu2011} (with dependence on
$n$ and $\delta$ unchanged)
\end{remark}

\begin{remark}
We note the somewhat surprising fact that the bound in~\ref{thm:main}
does not depend explicitly on $p$.
This is a result of the presence of
regularization parameter $r$, which prevents
$p\calK + r$ from becoming too sparse.
We note that if one imposes stronger assumptions on the
growth of $p$ (namely, restricting the speed with which $p$ can
approach $0$),
our proofs can be adapted to dispense with $r$ altogether,
in which case $p$ appears in the bounds instead.
\end{remark}

Our main tool for proving Theorem \ref{thm:main} is the
Davis-Kahan theorem~\cite{DavKah1970},
which we use in the form presented in~\cite{RohChaYu2011}.
We here index all quantities by $n$ to reiterate that
all quantities are allowed to depend on $n$, but remind the reader
that we will drop this indexing in much of the sequel for
ease of notation.
\begin{theorem} \label{thm:RCYDK}
Let $S_n \subset \R$ be an interval and
let $\calX_n$ be a matrix with orthonormal columns that span
the same subspace as that spanned by the eigenvectors of
$\calL^2(p_n \calK^{(n)})$
with corresponding eigenvalues in
$$ \lambda_{S_n}(\calL^2(p_n \calK^{(n)} + r_n J))
	= S_n \cap \lambda(\calL^2(p_n \calK^{(n)} + r_n J)).$$
Define $X_n$ analogously for $\calL^2(Y^{(n)} + r_n J)$.
Let $\delta_n$ be defined for $\calL^2(p_n \calK^{(n)} + r_n J)$
as in~\eqref{eq:deltadef}.

If $\calX_n$ and $X_n$ are of the same dimension,
then there exists orthonormal matrix $O_n$,
which depends on $\calX_n$ and $X_n$, such that
\begin{equation*} \begin{aligned}
\frac{1}{2} \| X_n & - \calX_n O_n \|_F^2 \\
  &\le \frac{ \| \calL^2(Y^{(n)} + r_n J)
	- \calL^2(p_n \calK^{(n)} + r_n J) \|_F^2 }{ \delta_n^2 }.
\end{aligned} \end{equation*}
\end{theorem}

To apply Theorem~\ref{thm:RCYDK} toward Theorem~\ref{thm:main},
we need a concentration bound for $\calL^2(Y + rJ)$
about $\calL^2(p\calK + rJ)$.
We note that $Y,$ $\calK$, $J$ and $r$ all implicitly depend on $n$,
a fact that we do not generally make explicit in the sequel
for ease of notation,
but which we highlight here for clarity.
For each $n=1,2,\dots$, let $\calK^{(n)}$ be a weighted adjacency matrix
for a graph on $n$ points in $\bcalX$ as defined in~\eqref{eq:kernel}.
Similarly, let $Y^{(n)}$ be the corresponding sparse noisy kernel
matrix as defined in~\eqref{eq:model}.

\begin{figure}[ht]
  \centering
    \includegraphics[width=0.9\columnwidth]{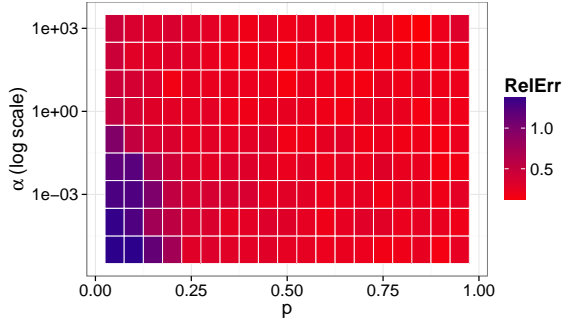}
  \caption{ Relative error (RelErr) in recovering the clean embedding
        of the high-dimensional swiss roll
	as a function of noise and occlusion.
	Each tile reflects the mean of 50 independent trials.
	We see that recovery is possible with low relative error except
	in the extreme case of simultaneous high-noise
	and heavy occlusion, suggesting that the embeddings are robust
	to both noise and occlusion of the kernel matrix. }
  \label{fig:HDS_ErrByNoiseAndOcc}
\end{figure}

\begin{theorem} \label{thm:Lbound}
Assume that regularization parameter $r$ grows with $n$ in such a way
that $r = \omega(n^{-1} \log n)$.
There exist constants $C,c > 0$ such that for suitably large $n$,
$$ \| \calL^2(Y+rJ) - \calL^2(p\calK + rJ) \|_F
	\le C \frac{ \log^{1/2} n }{ r n^{1/2} } $$
with probability at least $1 - n^{-c}$.
\end{theorem}
\begin{IEEEproof}
This theorem is proven in the Appendix.
\end{IEEEproof}

\begin{remark}
A number of results exist concerning concentration of the adjacency
matrix and the graph Laplacian of random graphs (see, for
example,~\cite{FeiOfe2005,Oliviera2010,RohChaYu2011,Tropp2012,LeLevVer2015,LeVer2015}).
In general, these results show that the graph Laplacian concentrates
in spectral norm about its mean when the quantity
$d = n\max_{1 \le i < j \le n} p_{ij}$ is of size $\Omega( \log n)$
(here $p_{ij}$ is the probability of an edge appearing between
nodes $i$ and $j$ in the random graph).
Our result differs from most of these,
in that we are concerned with concentration under the
Frobenius norm, rather than the spectral norm.
We obtain results in a similar regime,
as captured by our lower bound requirements on the
regularization term $r$.
\end{remark}

A key quantity in Theorem~\ref{thm:RCYDK}
is the spectral gap $\delta_n$ as defined in~\eqref{eq:deltadef}.
$\delta_n$ measures how well the eigenvalues
in $\lambda_S(\calL^2(p\calK^{(n)}))$
are isolated from the rest of the spectrum.
$\delta_n$ must grow in such a way that for suitably large $n$,
the eigenvalues falling in $S_n$ correspond to
the eigenvectors of interest, and the rate of this growth
is one of the factors controlling the convergence
in Theorem~\ref{thm:main}.
The existence of this eigengap is crucial for the application of
the Davis-Kahan Theorem~\cite{DavKah1970,RohChaYu2011}.
The eigengap depends on the matrix $p \calK^{(n)}$
(i.e., on the topology of the graph this matrix encodes).
As discussed in~\cite{Luxburg2007}, the existence of such a gap
is a reasonable assumption when, for example,
the data set (viewed through similarity function $\kappa$)
has a cluster structure.

\begin{figure}[ht]
  \centering
    \includegraphics[width=\columnwidth]{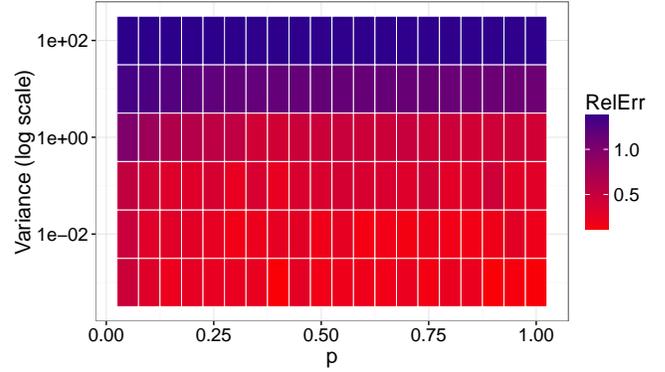}
  \caption{ Relative error in recovering the Laplacian
        eigenmaps embedding of the high-dimensional swiss roll
        as a function of occlusion and variance $\nu^2$
	in the multiplicative error model
	described in Equation~\eqref{eq:nonlinearnoise}.
        Each tile is the mean of 50 independent trials.
        We see that Laplacian eigenmaps is robust to
	moderate amounts of multiplicative noise, with reasonably good
	recovery at all values of $p$ provided $\nu^2 \le 1$
	(which we recall is five times the kernel bandwidth $\sigma=0.2$),
	but performance degrades sharply when uncertainty on the
	distance measure becomes too large ($\nu^2 \ge 10$).}
  \label{fig:HDSmult}
\end{figure}

\begin{figure*}[ht]
  \centering
  \subfloat[]{ \includegraphics[width=0.5\columnwidth]{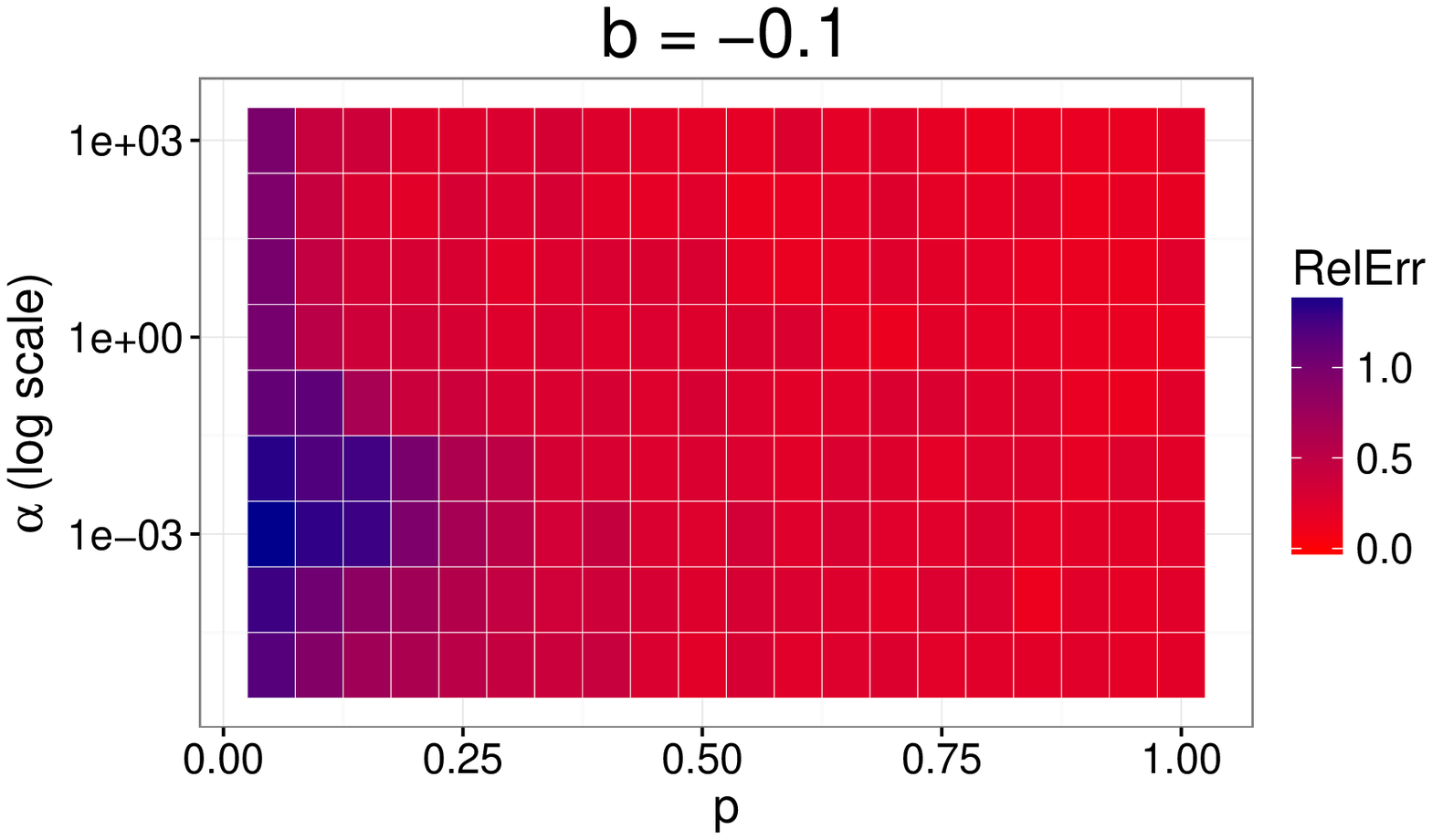} }
  \subfloat[]{ \includegraphics[width=0.5\columnwidth]{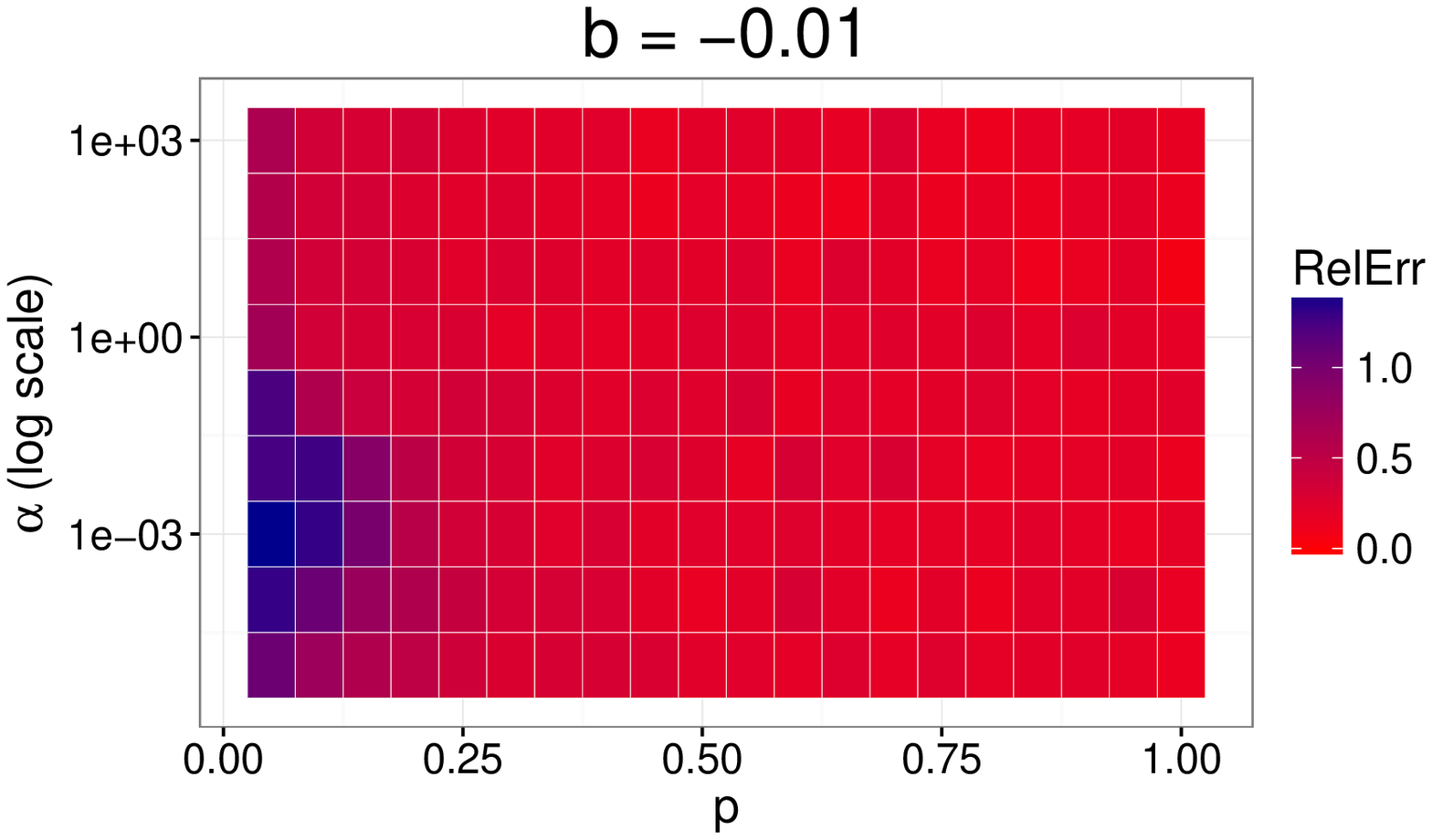} }
  \subfloat[]{ \includegraphics[width=0.5\columnwidth]{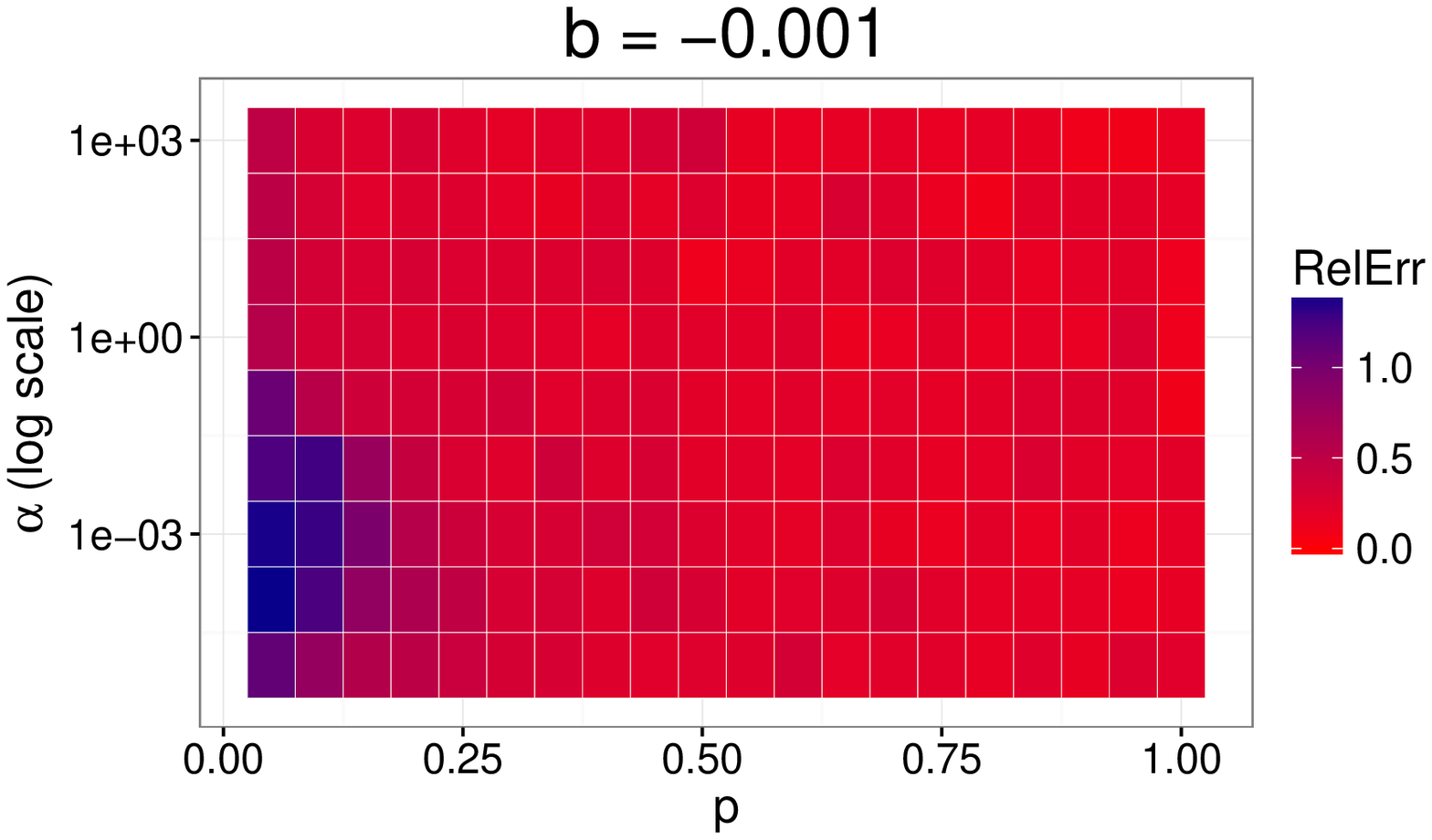} }
  \subfloat[]{ \includegraphics[width=0.5\columnwidth]{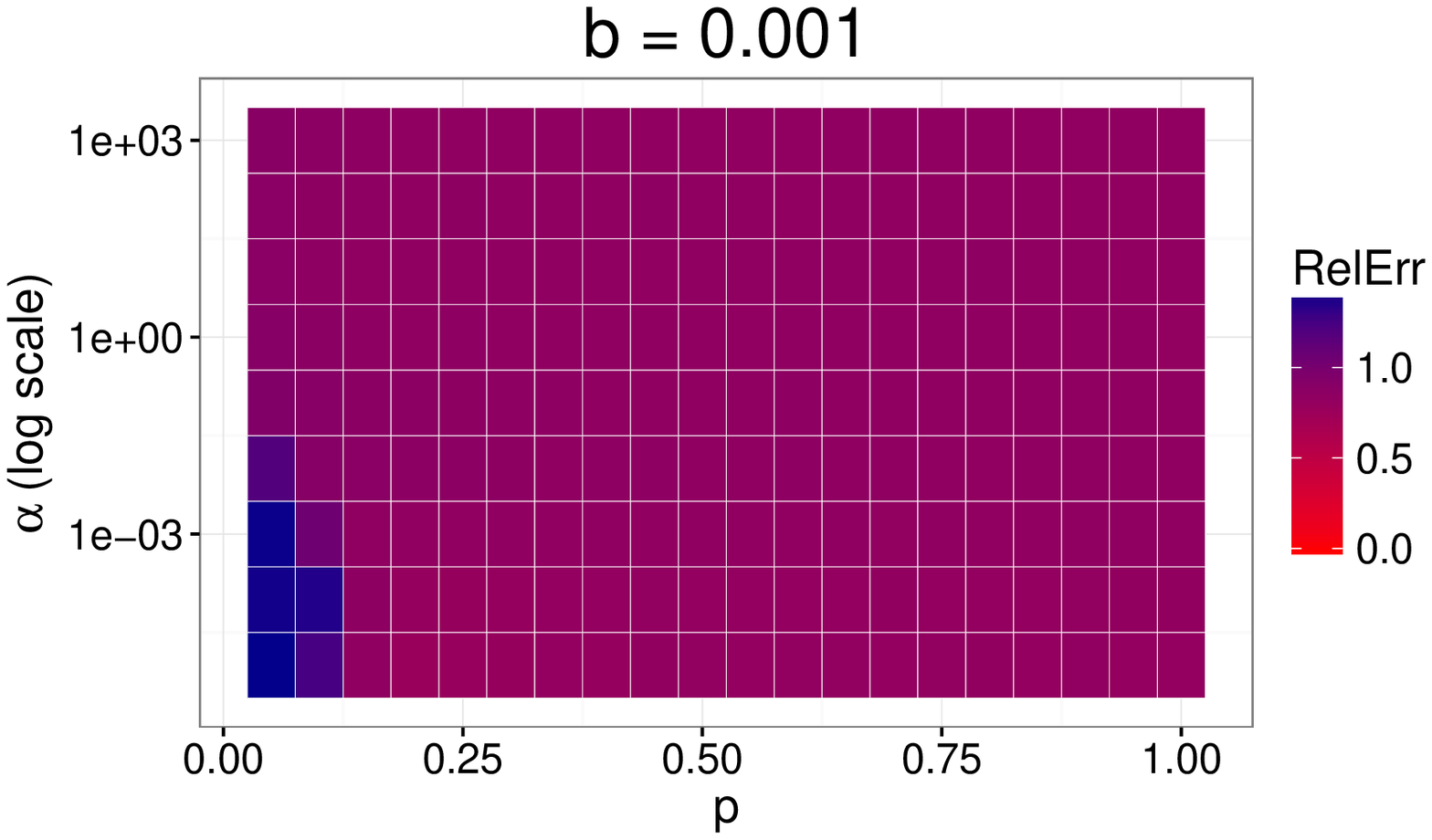} }
  \vspace{-0.2cm}
  \caption{ Relative error (RelErr) in recovering the clean embedding
	of the high-dimensional swiss roll
        as a function of occlusion and noise level for
        different levels of bias $b$.
	Each tile is the mean of 50 independent trials.
	We see that Laplacian eigenmaps embedding is quite robust to
	negative bias, but that even a small amount of positive bias
	in the errors causes a marked decrease in performance at
	all noise and occlusion levels. }
  \label{fig:HDSbias}
\end{figure*}

Typically, computing the Laplacian eigenmaps embedding of a data set is
not an end in itself, but rather a processing step performed prior to
subsequent inference, classification, or data exploration.
Such tasks depend entirely upon the geometry of the embedded data points
produced by Laplacian eigenmaps.
If the geometry of the points produced from the inexpensive
embedding based on $Y$ is approximately equal (up to rotation)
to that of the embedding based on $\calK$,
then we can expect comparable performance on downstream
tasks that are invariant under rotations of the data (e.g., clustering).
Thus, our results show that we can obtain performance comparable to that
obtained when using the dense, computationally intensive $\calK$
while avoiding the expense of working with $\calK$ directly.

\section{Experiments}
\label{sec:experiments}
In this section, we present simulation and real-world data
to complement our theoretical
results in Section~\ref{sec:mainresults}.

\subsection{Data Sets}
\setcounter{paragraph}{0}
We consider three data sets, one synthetic, one from connectomics,
and one from the speech processing literature.

\begin{figure*}[ht]
  \centering
    \subfloat[]{ \includegraphics[width=0.9\columnwidth]{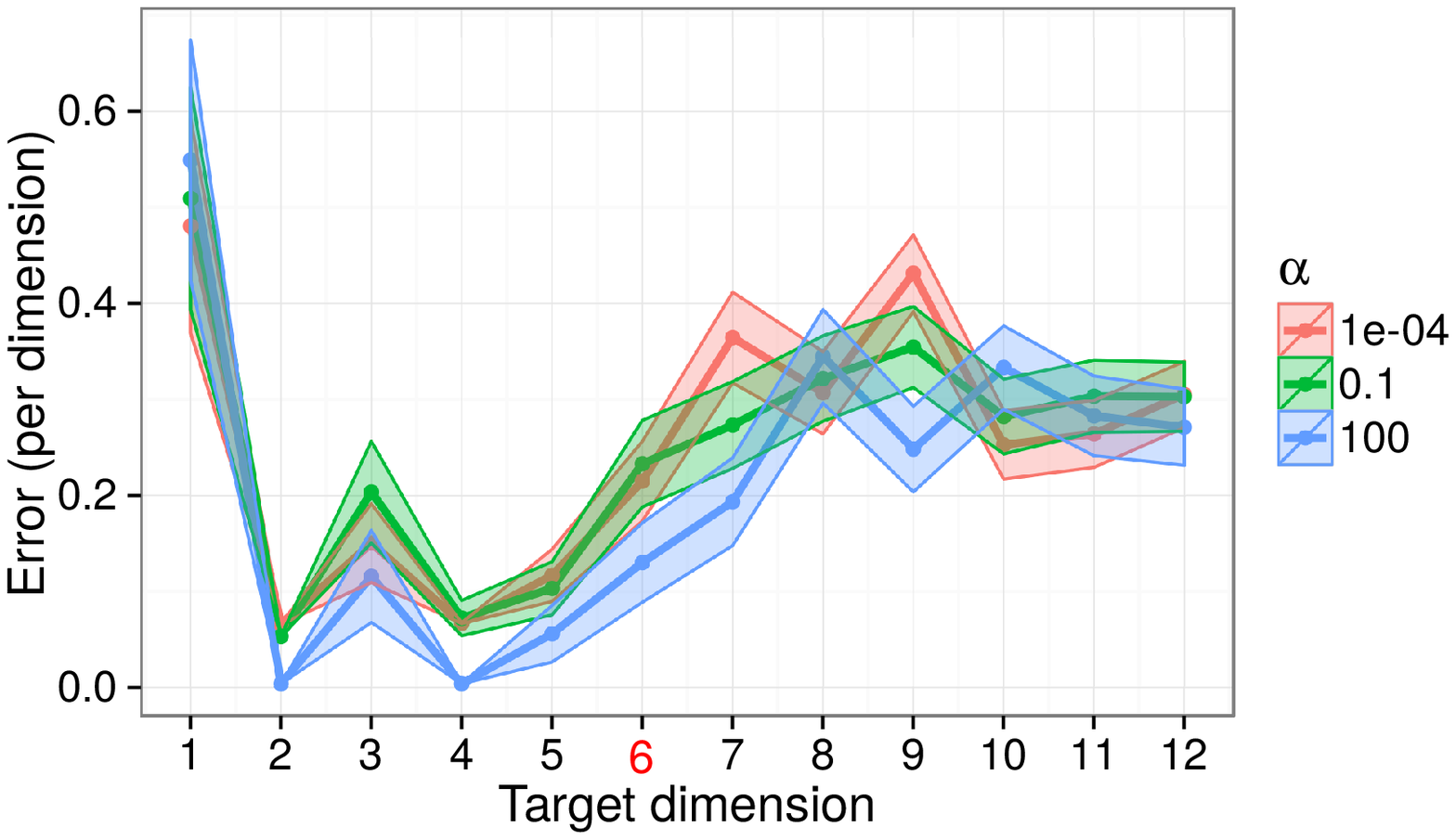} }
    \subfloat[]{ \includegraphics[width=0.9\columnwidth]{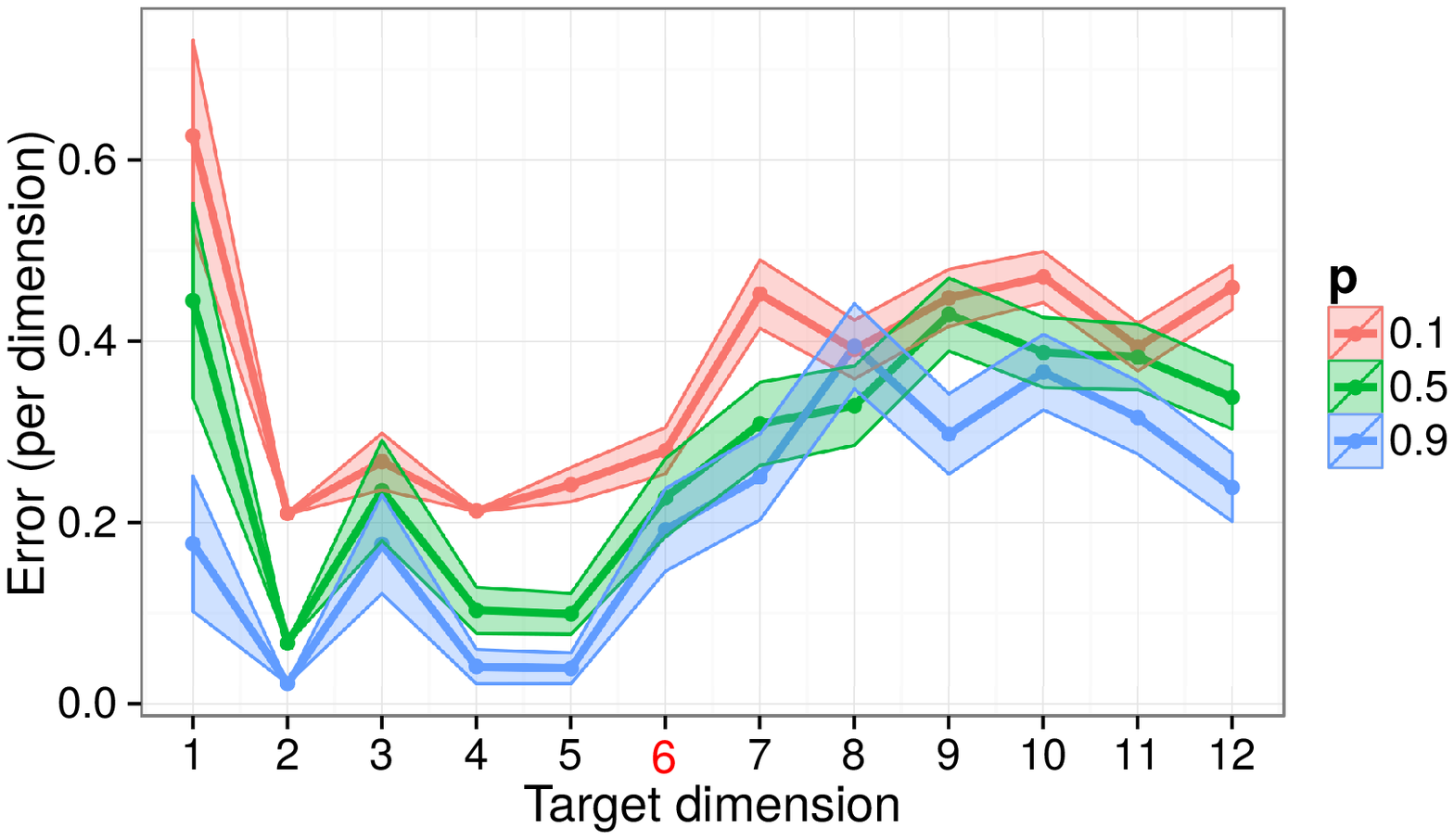} }
  \caption{ Relative error in recovering the Laplacian
        eigenmaps embedding of the high-dimensional swiss roll
        as a function of dimension at (a) different
        values of fidelity parameter $\alpha$
        and (b) different expected fractions of observed entries $p$ (right).
        The true underlying dimension of the data is highlighted in red.
        Each data point is the mean of 50 independent trials,
        with error bars indicating one standard error.
        We see a pattern typical of model selection problems, in which the
        expressiveness of the model (i.e., higher embedding dimension)
        comes at the cost of increased variance
        (i.e., higher relative error in recovering the clean embedding). }
  \label{fig:HDS_ErrByDim}
\end{figure*}

\paragraph{Synthetic Data (Fig.~\ref{fig:3Dswiss},~\ref{fig:HDS_ErrByNoiseAndOcc},~\ref{fig:HDS_ErrByDim},~\ref{fig:HDSmult},~\ref{fig:HDSbias})}

We consider a high-dimensional analogue of the 3-dimensional
swiss roll manifold (see Fig.~\ref{fig:3Dswiss}).
We sample $n$ points uniformly at random from the
$d^*$-dimensional unit cube and embed those points into
$(d^* + 1)$-dimensional space by applying the swiss roll transform
$$ (x,y) \mapsto (c x \cos( cx ), y, cx \sin( cx ) ),~~~
	x \in \R, y \in \R^{d^*-1}$$
where $c$ controls the curvature of the manifold.
In all experiments we use $n=5000$, $d^* = 6$ and $c=5$.
We chose this higher-dimensional version of the well-understood,
simple swiss roll manifold to examine the effect of both
under- and over-estimating the dimension $d^*$.
We obtain a kernel matrix $\calK$ from these points
by applying a Gaussian kernel with bandwidth $\sigma$.
Results are fairly stable for a wide range of values of $\sigma$.
We use $\sigma=0.2$ in all experiments, while stressing that the task of
selecting parameters in dimensionality reduction techniques warrants
much additional study.

\paragraph{{\it C. elegans} Connectome (Fig.~\ref{fig:celegans_ap_by_dim})}

We consider the task of clustering the 253 non-isolated neurons in the
{\it C. elegans}, a nematode commonly used as a simple biological model
(see~\cite{CheVogLyzPri2016} and citations therein).
These neurons are categorized according to
their function: sensory neurons, interneurons and motor neurons,
which make up 27.96\%, 29.75\% and 42.29\% of the connectome, respectively.
Our data consists of the symmetric binary
adjacency matrix corresponding to the {\it C. elegans} brain graph,
in which each node corresponds to an individual neuron,
with an edge between two neurons if they share a synapse.
As discussed in~\cite{CheVogLyzPri2016}, this brain graph can be constructed
in multiple ways.
Here we consider the subgraph of the chemical connectome induced by
the non-isolated vertices of the electrical gap junction connectome. 
Our goal is to embed the nodes of this graph via Laplacian eigenmaps
so that clustering (e.g., by $k$-means) recovers the three neuron
categories enumerated above.
We assess the quality of these embeddings using
adjusted Rand index (ARI)~\cite{HubAra1985}, which measures how well two
partitions agree, adjusted for chance.

\begin{figure}[ht]
  \centering
    \includegraphics[width=0.8\columnwidth]{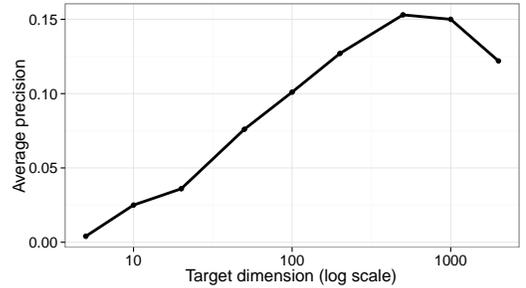}
  \caption{ Performance on the speech task, measured by average precision,
	as a function of embedding dimension.
	We see that performance peaks at an embedding dimension of $d=500$,
	with a severe degradation in the case where embedding dimension is
	chosen too small. }
  \label{fig:AP_by_dim}
\end{figure}

\paragraph{Speech Data (Fig.~\ref{fig:AP_by_dim},~\ref{fig:AP_by_NoiseAndOcc} and~\ref{fig:AP_by_NoiseAndOcc_Reg})}

We consider a speech processing data set
used in~\cite{LevHenJanLiv2013,LevJanVan2015},
consisting of $10,383$ spoken word examples,
representing $5,539$ distinct word types.
We refer the reader to~\cite{LevHenJanLiv2013} for technical details.
Using DTW alignment cost, we define a radial basis kernel
on the word examples to obtain a $10,383 \times 10,383$
kernel matrix that serves as our starting point for constructing embeddings.
The evaluation, developed in~\cite{CarThoJanHer2011},
assesses how well a representation distinguishes word types
as measured by average precision (AP),
which runs between $0$ and $1$, with $1$ representing perfect performance.
Performance on this task for this data set varies depends on many
factors, e.g., choice of acoustic features,
and better performance than reported here has been obtained.
However, the aim of this paper is not to best that performance, but rather
to examine how noise and occlusion influence performance
for a given set of observations.

\subsection{Noise Conditions}
\setcounter{paragraph}{0}
We consider the effects of additive noise and occlusion both in isolation
and in tandem on the quality of Laplacian eigenmaps embeddings.

\paragraph{Additive Noise}
Given a kernel matrix $\calK \in [0,1]^{n\times n}$, we produce a random
symmetric matrix $K \in [0,1]^{n \times n}$
where $K_{ii} = 0$ for all $i \in [n]$,
and $\{K_{ij} \}_{1\le i < j \le n}$
are independent with $K_{ij}$ beta-distributed with $\E K_{ij} = \calK_{ij}$.
We constrain the expected value of beta-distributed $K_{ij}$
in this way by fixing one of the two shape parameters of the beta
distribution, and varying the other to change the variance of the $K_{ij}$.
In particular,
$K_{ij} \sim \betadistrib(\alpha_{ij},\eta_{ij})$
with $\alpha_{ij} > 0$ and $\eta_{ij} > 0$.
fixing $\eta_{ij} = \alpha_{ij}(1-\calK_{ij})/\calK_{ij}$ ensures
that $\E K_{ij} = \calK_{ij}$ with
$$ \VAR K_{ij} = \frac{ \calK_{ij}^2 (1-\calK_{ij}) }
			{ \alpha_{ij} + \calK_{ij} } , $$
so that we can vary our level of uncertainty on the $K_{ij}$ variables
by varying $\alpha_{ij}$.
We select a single global value $\alpha > 0$, and take
$K_{ij} \sim \betadistrib(\alpha, \alpha(1-\calK_{ij})/\calK_{ij}).$
In the limit $\alpha \rightarrow 0$, the $K_{ij}$ are simply Bernoulli
random variables with probability of success $p_{ij} = \calK_{ij}$.
In the limit $\alpha \rightarrow \infty$, we have $K_{ij} = \calK_{ij}$
almost surely.
Thus, we can think of our parameter $\alpha$ as a measure of the accuracy
of our measurements of $\calK$.
We note also that our parameterization implies that the $K_{ij}$ variables
do not all have the same variance.
Rather, variances are smaller for $\calK_{ij}$ nearer to $0$ and $1$.
As discussed in Section~\ref{sec:motivation}, this is a good model for
applications in which the cases $\calK_{ij} \approx 0$ and
$\calK_{ij} \approx 1$ are comparatively easy to handle
from an estimation or computation standpoint,
and the trouble arises from the cases where $\calK_{ij} \approx 1/2$.

\paragraph{Occlusion}
We observe an occluded version of $\calK$, where entries above the diagonal
are observed independently with probability $p$.
We proceed with our embedding using this sparse kernel matrix,
with zeros in the unobserved entries.

\paragraph{Additive Noise with Occlusion}
This condition combines the preceding two. We observe an occluded,
noisy version of matrix $\calK$. That is, we generate noisy matrix $K$
from $\calK$ with entries drawn independently from suitably chosen
beta-distributions, then occlude $K$ by independently observing
entries with probability $p$.

\paragraph{Multiplicative and Biased Errors with Occlusion}
Rather than the unbiased additive noise considered above,
we consider how more complicated multiplicative and biased errors
influence the quality of Laplacian eigenmaps embeddings.
As discussed in Section~\ref{sec:motivation}, provided these errors are
sufficiently well-behaved, we can adapt the results presented in this paper
to make similar statements about this more general error model.

\subsection{Effect of Noise and Occlusion on Embeddings}
\setcounter{paragraph}{0}

Our main theoretical result suggests that Laplacian eigenmaps embeddings
should be robust to noise and occlusion.
Fig.~\ref{fig:HDS_ErrByNoiseAndOcc} shows how noise and occlusion
influence the error in recovering the clean Laplacian eigenmaps embedding.
Here, the target dimension is fixed at $d=d^*=6$,
while the noise and occlusion vary on the two axes.
Each tile is the relative error averaged over $50$ independent trials.
We see that the clean Laplacian eigenmaps embedding is recovered with
low error over a wide range of noise levels and occlusion
rates, with performance degrading only when the fraction of
observed entries goes below $0.25$ in high-noise conditions.

\begin{figure*}[ht]
  \centering
    \includegraphics[width=0.66\columnwidth]{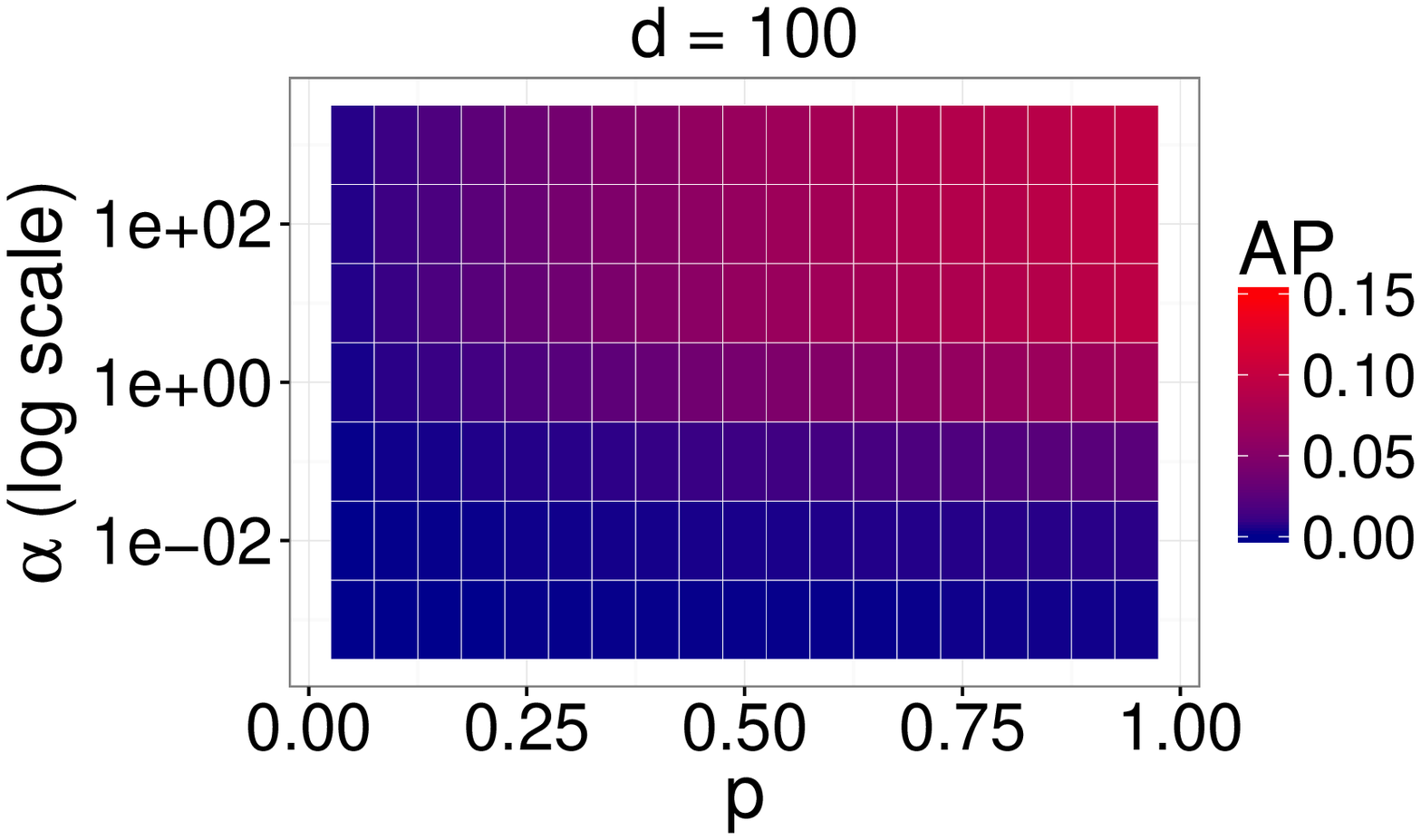} 
    \includegraphics[width=0.66\columnwidth]{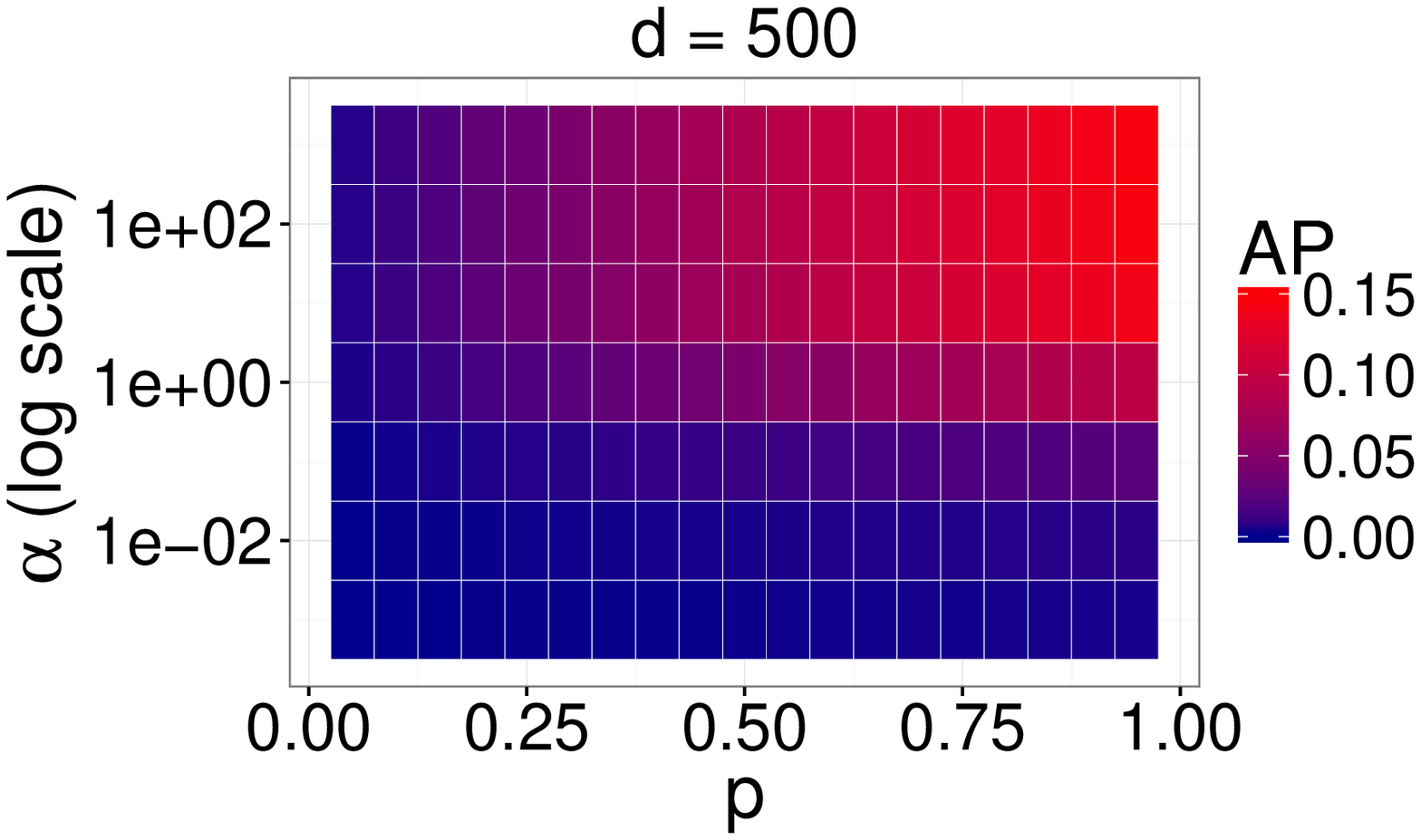} 
    \includegraphics[width=0.66\columnwidth]{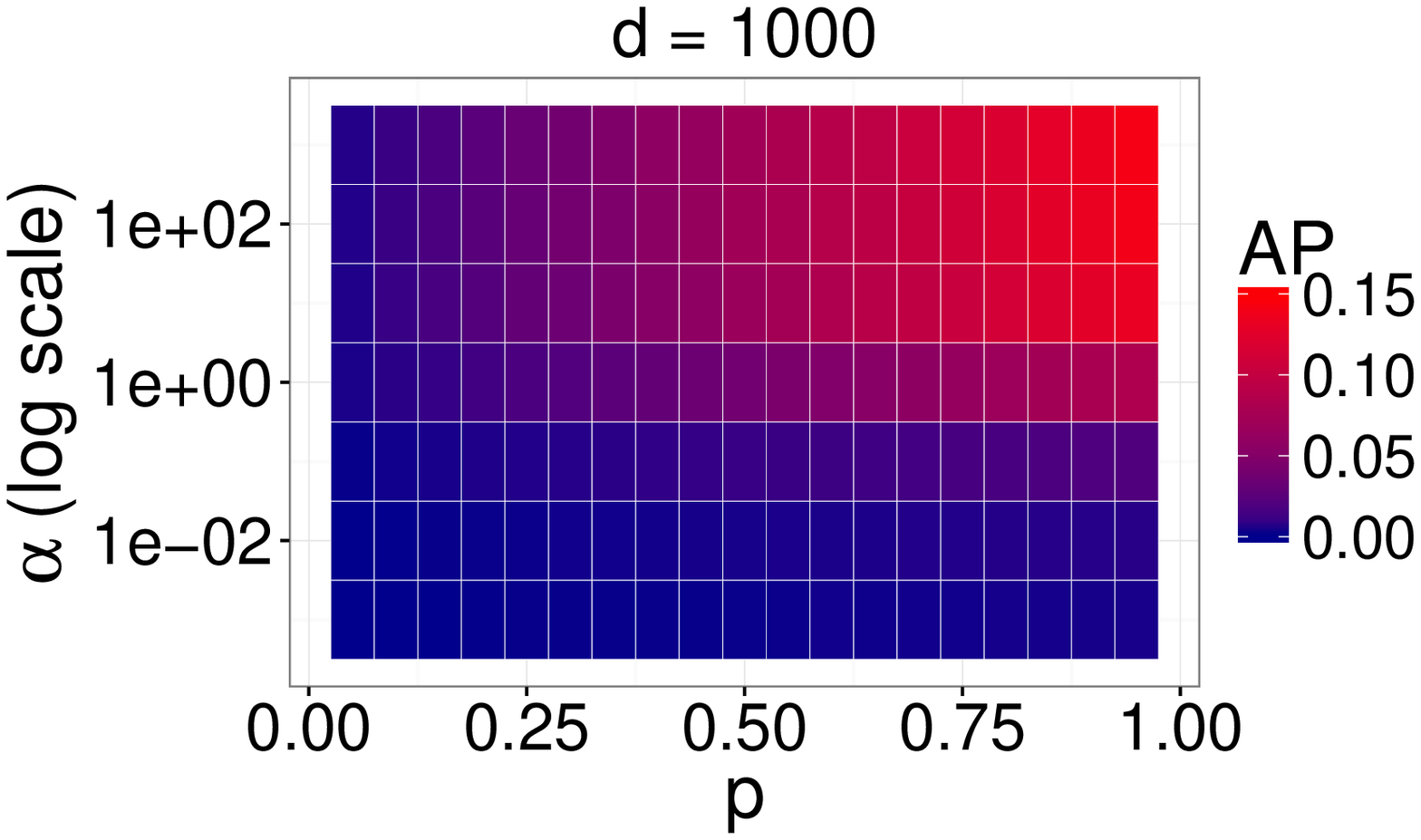} \\
  \vspace{-0.25cm}
  \caption{Average precision (AP) on the speech data set
	as a function of occlusion and noise level for
	different embedding dimensions $d$. Each tile is the mean of ten
	independent trials. We see that performance degrades similarly
	for all three target dimensions
	in the presence of noise and occlusion. }
  \label{fig:AP_by_NoiseAndOcc}
  \vspace{-0.25cm}
\end{figure*}

Fig.~\ref{fig:AP_by_NoiseAndOcc} further illuminates the results seen
in the synthetic data.
Rather than looking at the relative
error in recovering the clean embedding,
we examine how noise and occlusion in the kernel matrix
influence the down-stream speech task of distinguishing word
types. The plot shows average precision as a function of both noise level
and occlusion for three different embedding dimensions.
We see that performance decays similarly in all three embedding dimensions,
but that choice of embedding dimension has a large effect on
overall performance. For example, comparing the $d=100$ case
with the $d=500$ case, we see that both exhibit similar
deterioration patterns with respect to noise level and
expected fraction of observed entries,
but the 500-dimensional embeddings out-perform
the 100-dimensional ones when noise and occlusion are not so severe as to
drown out the signal in the kernel matrix.

\subsection{Effect of Multiplicative Error and Bias}

Our theoretical results are for the case of unbiased noise,
$\E K_{ij} = \calK_{ij}$,
and it is natural to ask whether similar results hold for
a broader class of error models.
As mentioned in Section~\ref{sec:motivation}, our results can be
extended to biased errors ($\E K_{ij} \neq \calK_{ij}$),
provided those errors are suitably well-behaved.
Fig.~\ref{fig:HDSmult} and~\ref{fig:HDSbias} lend experimental
support to this point.

Using the same synthetic high-dimensional swiss roll
setup as in Fig.~\ref{fig:HDS_ErrByNoiseAndOcc},
we consider biased noise,
with $K_{ij}$ beta distributed, but with
$\E K_{ij} = \calK_{ij} + b$, where $b \in \R$ is a bias,
clipping $\calK_{ij} + b$ to lie in $[0,1]$ in the event that
the bias $b$ pushes $\calK_{ij}$ out of its allowed range.
Note that this corresponds to making $K_{ij}$ either identically $0$
or identically $1$, according to whether $\calK_{ij} + b$ is
less than $0$ or greater than $1$, respectively.
We again vary the parameter $\alpha$ as described above,
but now the errors are biased away from $\calK_{ij}$.
Fig.~\ref{fig:HDSbias}
shows relative error in recovering the clean embeddings,
again as a function of the parameters $p$ and $\alpha$,
for four different levels of bias $b = -0.1,-0.01,-0.001,0.001$.
The first thing we notice is that performance is far more sensitive
positive bias than it is to negative bias,
with negative bias as large as $-0.1$ (a full one tenth of the dynamic
range of the similarity measure)
having comparatively little effect while a positive bias of just
$0.001$ results in notably worse relative error at all
levels of noise and occlusion when compared to the unbiased errors
in Fig.~\ref{fig:HDS_ErrByNoiseAndOcc}.
This performance makes sense.
Positive bias in our estimation of $\calK$ results in us embedding a graph
that looks highly connected, and the signal present in the comparatively
sparse $\calK$ is swamped.
On the other hand, negative bias in our estimates only serves to further
accentuate the few high-weighted observed entries,
since only those entries for which $\calK_{ij}$ is suitably far from
$0$ survive the bias.
We have observed empirically that a similarly-motivated technique,
in which small entries of the kernel matrix are clipped to $0$,
yields slight performance improvements in speech applications.

We further explore how general errors influence the
quality of Laplacian eigenmaps embeddings by
considering an error model in which
\begin{equation} \label{eq:nonlinearnoise}
  K_{ij} = \exp\{ -D_{ij}^2/\sigma^2 \},
\end{equation}
where $D_{ij} = d(x_i,x_j) + Z_{ij}$, and $Z_{ij}$ is a one-dimensional
normal random variable with mean $0$ and variance $\nu^2$.
Thus, we have a distance measure corrupted by unbiased noise,
corresponding to the common scenario in which the kernel function
$\kappa(x,y)$ is a function of the distance between objects $x$ and $y$
and uncertainty lies in the measurement of that distance.
The result, in the case of a nonlinear kernel function,
is (typically) non-additive, biased, error,
so that $\E K_{ij} \neq \calK_{ij} = \kappa(x_i,x_j)$.
We again use the same high-dimensional swiss roll
as described above.
We generate noisy versions of the kernel matrix $\calK$,
using the same Gaussian kernel with bandwidth $\sigma=0.2$,
but now noise takes the form described in Equation~\ref{eq:nonlinearnoise}.
Fig.~\ref{fig:HDSmult} shows relative error in our recovery of the
clean embeddings, as a function of the fraction of observed entries $p$
and the variance $\nu^2$ of the noise term $Z_{ij}$.
We see that Laplacian eigenmaps embeddings are robust to fairly large
amounts of uncertainty in the distance measurement.
Indeed, we see that relative error is near zero for variance
$\nu^2 \le 1$, with the exception of particularly small values of $p$,
when nearly all of the kernel matrix is occluded.
This performance is impressive in light of the fact that $\nu^2 = 1$
corresponds to a standard deviation a full five times larger than the kernel
bandwidth in these experiments.

\subsection{Model Misspecification}
\setcounter{paragraph}{0}

\begin{figure*}[ht]
  \centering
    \includegraphics[width=0.66\columnwidth]{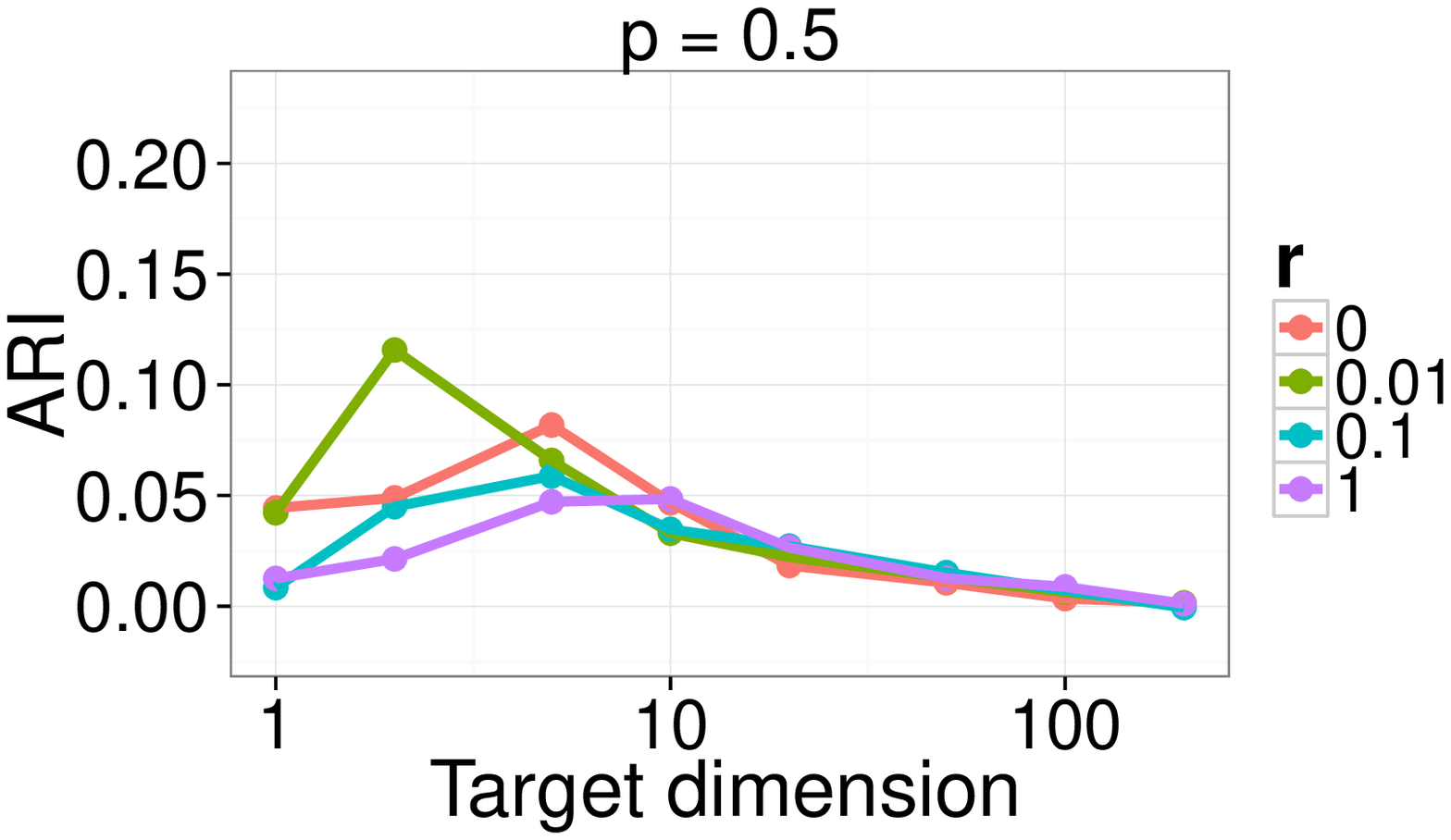} 
    \includegraphics[width=0.66\columnwidth]{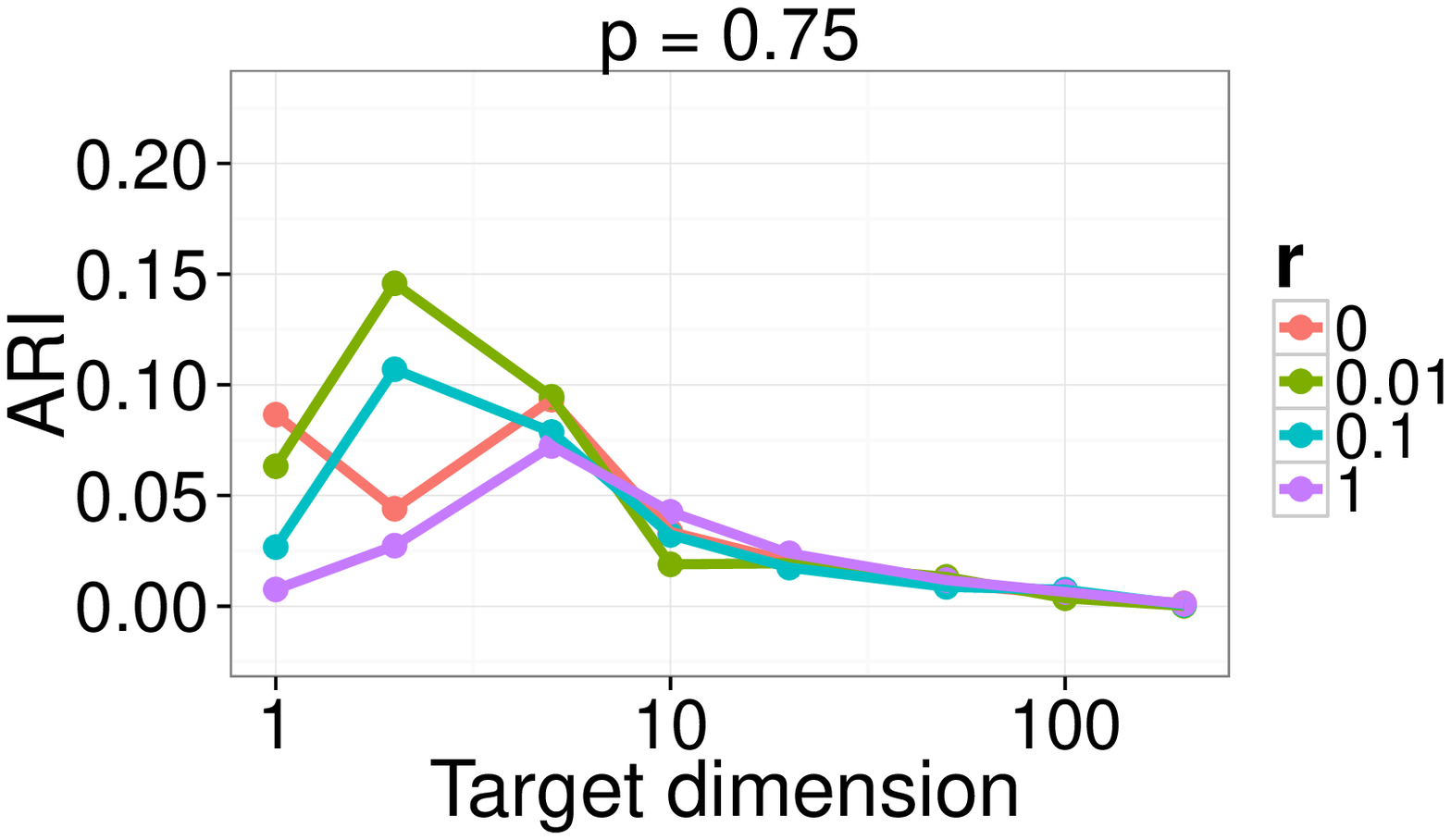} 
    \includegraphics[width=0.66\columnwidth]{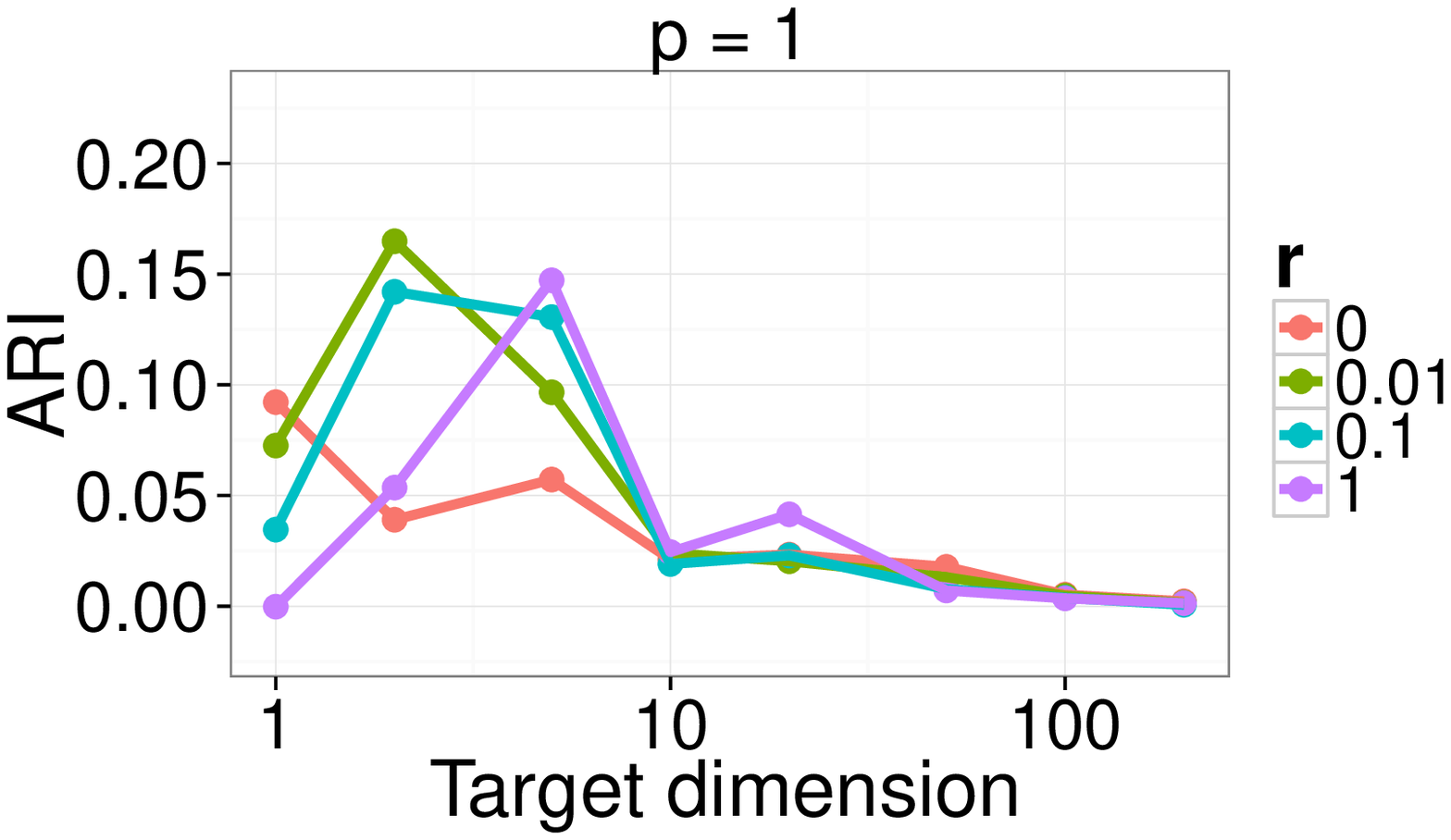} \\
  \vspace{-0.25cm}
  \caption{Adjusted Rand index (ARI) on the {\it C. elegans} data set
	for different levels of regularization
	as a function of dimension
	at different values of $p$,
	the expected fraction of observed entries.
	Each data point is the mean of 50 independent trials.
	We see that regularization enables us to accurately
	cluster the neurons even when much of the structure of the
	brain graph is occluded, with performance consitently
	superior to that obtained without regularization. }
  \label{fig:celegans_ap_by_dim}
  \vspace{-0.25cm}
\end{figure*}

Selecting the target dimension is of the utmost importance for
good embeddings.
Fig.~\ref{fig:HDS_ErrByDim} shows how embedding dimension interacts with
noise and occlusion on the synthetic data.
The two plots show that relative error in recovering the clean embedding
is smaller at lower target dimensionalities,
and this pattern holds over a wide range of noise levels and occlusion rates.
In particular, we note that relative error in the presence of
high noise and high occlusion remains comparable to the relative error
in low noise and low occlusion conditions.
Of course, this only tells part of the story.
Fig.~\ref{fig:AP_by_dim} shows average precision on the speech data set
under clean conditions, as a function of embedding dimension.
While a low-dimensional embedding performed under noise or occlusion
might very closely resemble the corresponding clean embedding
as in Fig.~\ref{fig:HDS_ErrByDim},
Fig.~\ref{fig:AP_by_dim} suggests that such an embedding would not yield
satisfactory performance on downstream tasks such as classification.
Indeed, we see here a pattern typical of model selection tasks:
one must balance estimation error of model parameters against
error in fitting the observed data~\cite{Shibata1986,FraRaf2002,RafDea2006}.
The noisy embedding can only be as good as the clean embedding we are
attempting to recover.

\subsection{Effect of regularization}

In the setting of the current work,
when $p$ is too small, we are in the sparse graph
setting~\cite{ChaChuTsi2012,AmiChaBicLev2013,JosYu2013,QinRoh2013,LeLevVer2015,LeVer2015},
and it is natural to consider whether applying regularization might ease
the deterioration of embedding quality in this regime.
We follow the regularization procedure described
in~\cite{LeLevVer2015}, in which a regularization parameter $r$ is added
to each entry of the observed matrix.
That is, letting $Y$ denote the occluded version of the noisy matrix $K$,
we apply Laplacian eigenmaps to the matrix $[Y_{ij} + r]$
rather than $Y$ itself.
Our main theoretical results suggest that under suitable conditions,
such an approach will be beneficial.
The {\it C. elegans} brain graph is extremely sparse,
and occlusion makes this sparsity still more dramatic.
Fig.~\ref{fig:celegans_ap_by_dim} shows how regularization influences
downstream performance on the {\it C. elegans} data under different levels
of occlusion.
We see that when $r$ is chosen too small, regularization is not enough to
significantly change the learned embedding.
Similarly, when $r$ is chosen too large,
regularization overpowers the signal present in the occluded matrix.
However, with the {\it C. elegans} data, we see that there exists a level
($r \approx 0.01$) at which regularization greatly improves ARI,
even when only half of the edges of the graph are known.
We note that embeddings produced by the regularization procedure described
in~\cite{QinRoh2013} resulted in nearly identical performance.

The performance seen here is especially exciting from the
neuroscience standpoint-- these results suggest that we can recover structural
and functional information in connectome data even when accurate assessment
of all possible neural connections is impossible.
We note the similarity of this phenomenon
to that explored in~\cite{PriSusTanVog2014},
where the authors considered graph inference in the setting where one can
trade the accuracy of edge assessment against the number of edges assessed.
Of course, the usefulness of this result requires that
can determine an appropriate value for $r$ for a given data set,
a problem that we leave for future work.

\begin{figure*}[ht]
  \centering
  \subfloat[]{ \includegraphics[width=0.5\columnwidth,height=1in]{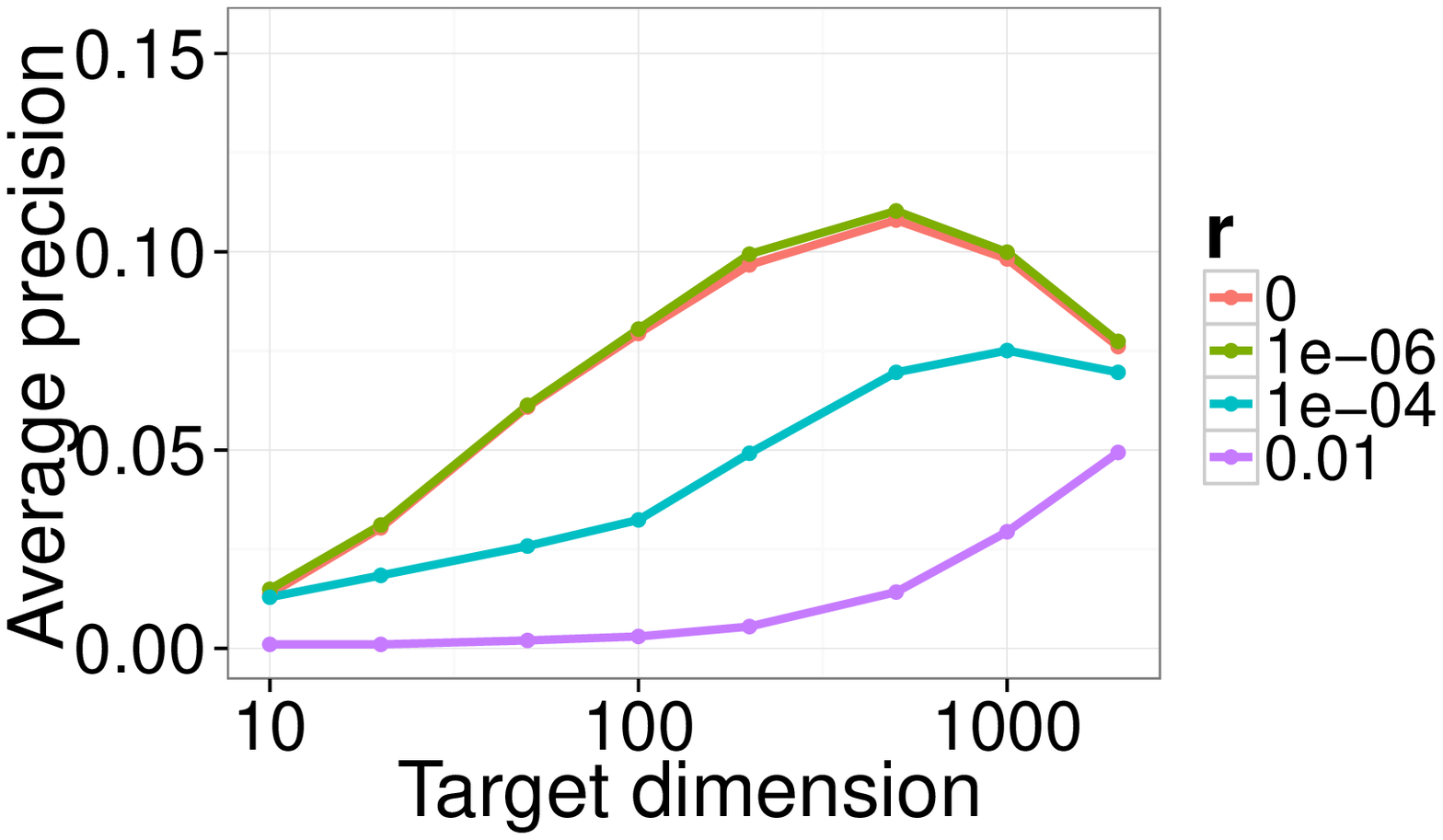} }
  \subfloat[]{ \includegraphics[width=0.5\columnwidth,height=1in]{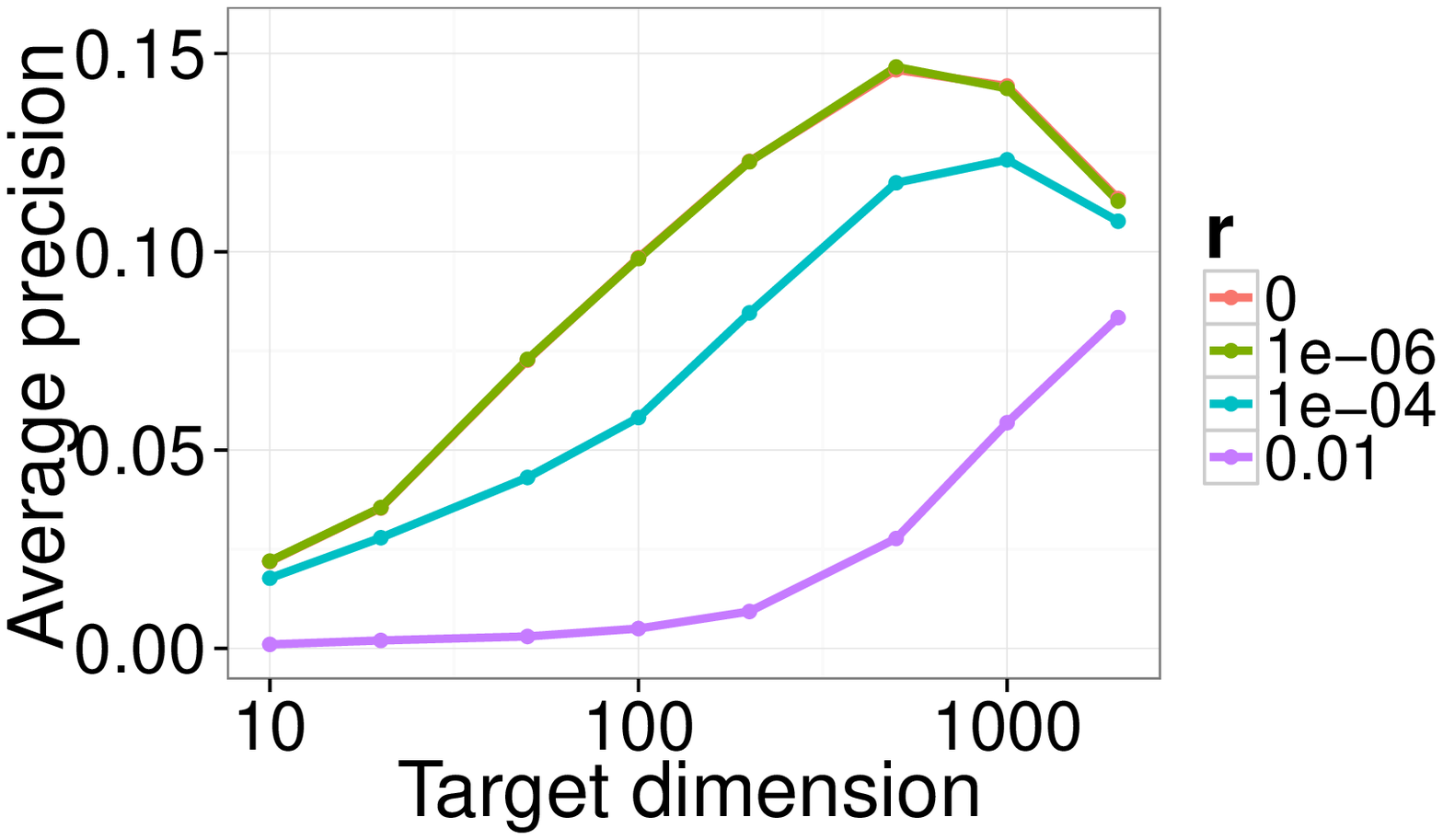} }
  \subfloat[]{ \includegraphics[width=0.5\columnwidth,height=1in]{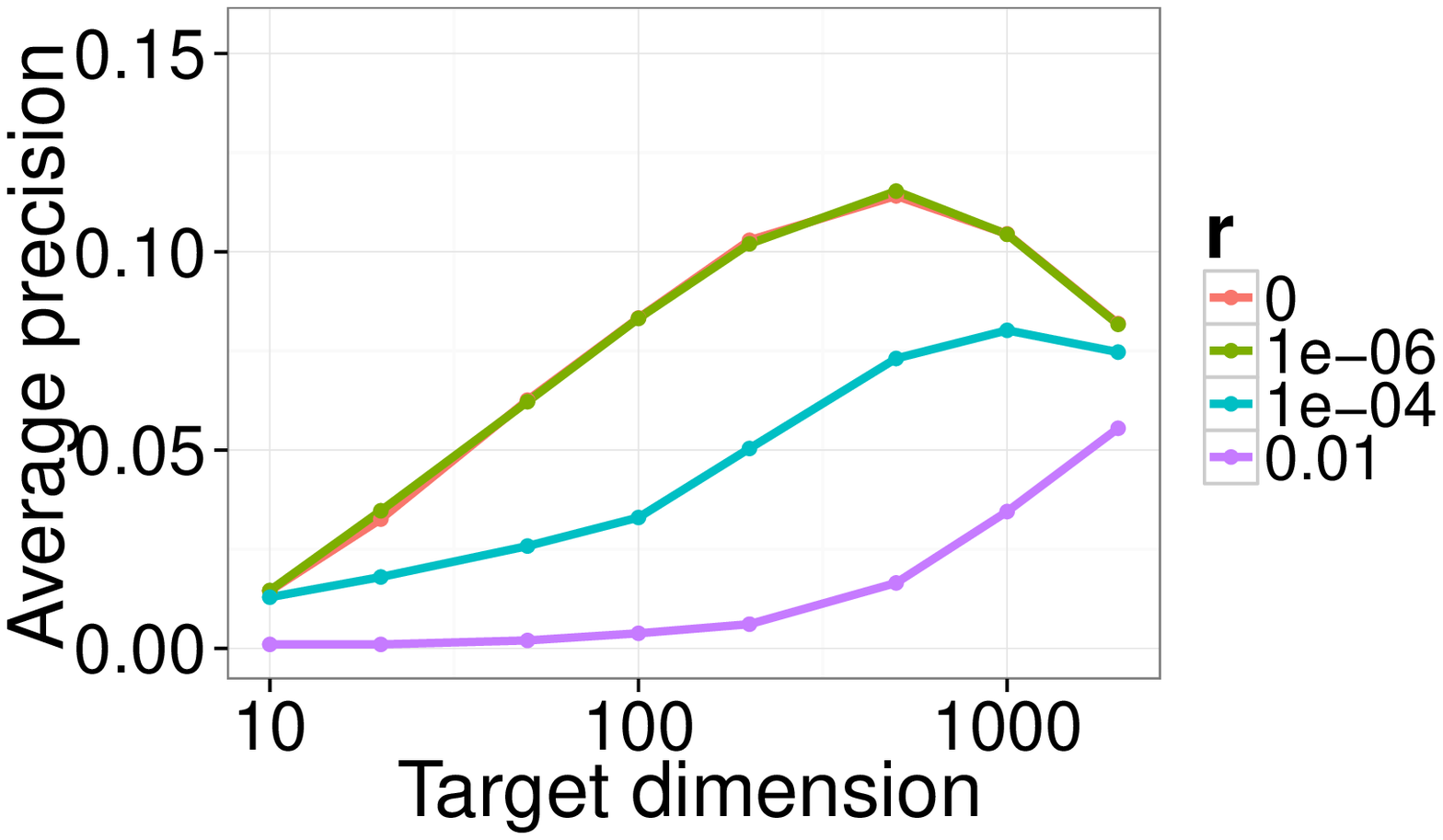} }
  \subfloat[]{ \includegraphics[width=0.5\columnwidth,height=1in]{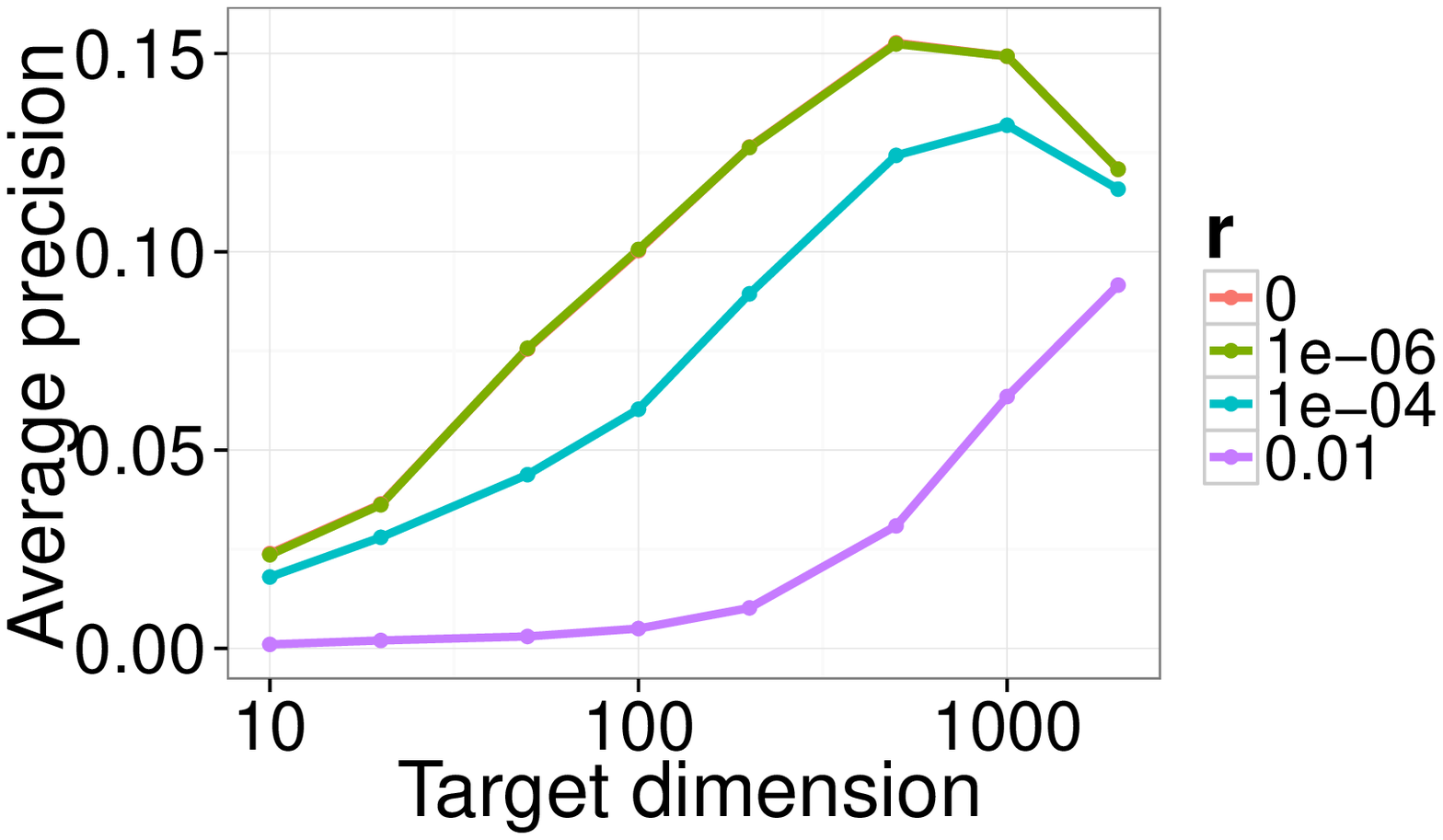} }
  \vspace{-0.2cm}
  \caption{ Average precision on the speech data set
       as a function of embedding dimension for different levels of
       regularization under varying amounts of noise and occlusion:
       (a) $\alpha = 10, p = 0.7,$
       (b) $\alpha = 10, p = 1.0,$
       (c) $\alpha = 100, p = 0.7,$
       (d) $\alpha = 100, p = 1.0.$
       Each data point is the mean of 10 independent trials.
       We see that while regularization does not provide the stunning
       improvement that it does on the {\it C. elegans} graph,
       moderate regularization at least does not noticeably harm
       average precision. }
  \label{fig:AP_by_NoiseAndOcc_Reg}
  \vspace{-0.25cm}
\end{figure*}

We close by illustrating conditions under which regularization does not
appear to be a benefit.
One would think, initially, and especially given the improvement seen
in the {\it C. elegans} data, that regularization would yield similar gains
in our speech task.
Fig.~\ref{fig:AP_by_NoiseAndOcc_Reg} shows how regularization
influences downstream performance on the speech task.
We see that regularization does not appear to confer the benefit seen in
the {\it C. elegans} data.
Crucially, however, moderate amounts of regularization do not
appear have any adverse effects on average precision.
One possible explanation for this phenomenon comes from the fact
that the kernel bandwidth used in~\cite{LevHenJanLiv2013} was chosen
so as to give the best possible average precision on precisely the task
we are using for evaluation. That is, since the kernel bandwidth has already
been tuned so as to yield high-quality embeddings, regularization can do
little to improve the embeddings.
But this explanation does not account for the fact that regularization
does not appear to confer any protection against occlusion and noise
in the kernel matrix. It is possible that the speech data set is such that
the kernel matrix is sparse enough that regularization does nothing to
pull us toward a better embedding. We leave further exploration of this
phenomenon to future work.

\section{Discussion}
\label{sec:discussion}

We have presented an analysis of the concentration of the graph Laplacian
of certain kernel matrices under occlusion and noise.
Crucial to our bound was the presence of a certain structure in the kernel
matrix that ensures concentration of the row-sums.
Experiments on both synthetic and real data show that a concentration
phenomenon similar to that predicted by the theory is present,
and has effects both on performance in downstream tasks and on
the model selection problem.
We close by briefly mentioning some directions for future work.

\subsection{Adaptive Techniques}
The regularization used here was applied uniformly to every vertex of the
graph, but regularization is only required to control the high
variance associated with small-degree nodes.
In light of this, one might consider regularization techniques that
apply only to nodes that require it.
It is unclear a priori whether such an approach would be advantageous,
since regularization 
does little to change the behavior of high-degree nodes.
However, it stands to reason that a well-designed adaptive technique might
enable convergence of the regularized estimate to the true expected graph,
rather than to its regularized counterpart as in the current work.
For example, if only a small fraction of the nodes in a given graph require
regularization, then the Frobenius error between the regularized
and non-regularized Laplacians can still go to zero even
if $r$ goes to zero slowly.

In a similar vein,
it stands to reason that a technique that evaluates entries of
the kernel matrix adaptively rather than the edge-independent occlusion
model considered here might achieve more accurate recovery of the
clean embeddings.

\subsection{Other Error Models}
The noise model we have considered is additive,
unbiased and entry-wise independent.
As discussed in Section~\ref{sec:motivation},
our results can be (approximately) extended to multiplicative,
biased noise models, at least for certain kernels.
However, the concentration bounds we have used
require a certain independence structure.
As such, it seems likely that novel techniques will be required
to handle entry-wise dependent noise and occlusion in the kernel matrix.
For example, the techniques in~\cite{ORoVuWan2016}
might be brought to bear,
except that they require structural assumptions
on $\calK$ that seem unlikely to hold for a non-linear kernel function.


\subsection{Graph Construction}
We have largely ignored the problem of constructing the $k$-NN or
$\epsilon$-graph, the first step in Laplacian eigenmaps
and spectral clustering.
Rather than using either of these constructions,
we have relied on the fact that the kernel matrix
can be made to resemble these graphs
by using, for example, a Gaussian kernel.
We believe that the our analysis can be extended to many of these
constructions simply by taking advantage of this resemblance.
We leave this extension for future work.

\subsection{Other Dimensionality Reduction Techniques}

To what extent are different embedding techniques robust to uncertainty
in similarity measures (as opposed to errors on the observations themselves)?
To the best of our knowledge, MDS and Laplacian eigenmaps
remain the only techniques for which such questions have been explored.
We believe that analyses similar to that pursued in
the current work should apply to
other dimensionality reduction techniques.
Indeed, given the results in~\cite{YanXuZhaZhaYanLin2007}, it would be
a surprise to learn that no such general result is possible.

As alluded to in Section~\ref{sec:relatedwork},
a natural approach to the problem considered in this paper would be
to apply Chatterjee's universal singular value thresholding
(USVT)~\cite{Chatterjee2015} to the occluded,
noisy kernel matrix $Y$
(or, in the case where $\kappa(x,y)$ is a function
of $d(x,y)$, to transform $Y$ into an occluded matrix of distances $D$,
impute the missing entries of $D$ using USVT,
and reapply the kernel function to obtain an estimate of $\calK$).
Applying USVT in this manner to the speech task
considered in Section~\ref{sec:experiments} yields results essentially
identical to those reported using $Y$ alone at all noise and occlusion rates.
Indeed, USVT performed remarkably similarly to our method
on all three data sets, a fact that warrants further exploration.

Some well-known dimensionality reduction techniques
can be adapted fairly easily to the model in
Equation~\eqref{eq:model} by using Chatterjee's USVT to impute the missing
entries of $Y$ and proceeding apace.
In an experimental setup identical to the synthetic high-dimensional
swiss roll experiments presented in Section~\ref{sec:experiments},
we explored the effect of noise and occlusion on
both MDS and kernel PCA (KPCA).
We found that neither of these methods compared favorably to the
results seen for Laplacian eigenmaps.
While direct comparison of the relative errors for these three different
methods is not possible (e.g., embeddings produced by MDS are not constrained
in the same way that Laplacian eigenmaps embeddings are),
from a qualitative standpoint,
MDS and KPCA both degraded much more severely in the presence of
noise and occlusion when compared with Fig.~\ref{fig:HDS_ErrByNoiseAndOcc}.
While a direct comparison (experimental or otherwise)
of Laplacian eigenmaps with other dimensionality
reduction techniques is not the focus of this paper,
a more thorough exploration of how different methods fare in the presence of
noise and occlusion (and how those methods might be adapted to lessen
the impact of uncertainty) warrants additional work in the future.

\section*{Acknowledgments}
The authors would like to thank
Carey E. Priebe and Minh Tang
for discussion in formulating the problem setup
and for their insightful feedback and discussion.

\appendix[Concentration of $\calL^2(Y + rJ)$]
\label{sec:apx}
In what follows, we suppress dependence on $n$ for ease of notation.
We remind the reader that all quantities involved, including
the parameters $r$ and $p$ all implicitly depend on $n$.
We let $\Yhat = Y + rJ$ denote the regularized
version of matrix $Y$, and define $\Dhat$ to be the corresponding
degree matrix, so that $\Dhat_{ii} = nr + \sum_{j=1}^n Y_{ij}$.
Denote the regularized version of $p\calK$ by
$\calKhat = p\calK + rJ$, with $\calDhat$ the corresponding
degree matrix, $\calDhat_{ii} = nr + \sum_{j=1}^n p\calK_{ij}$.

Throughout, $C > 0$ denotes a constant (independent of $n$),
which may change from line to line or from one lemma to another.
$\beta$ and $\gamma$ denote quantities
(both depending on $n$)
that will control convergence of the node degrees and
the Frobenius norm in Theorem~\ref{thm:Lbound}, respectively.
We will see that the constraints on $\beta$ and $\gamma$
required for our concentration bounds are such
that when we plug in $\gamma = C' n^{-1/2} r^{-1} \log^{1/2} n$
and $\beta = C'' n^{-1/2} r^{-1/2} \log^{1/2} n$
for suitably chosen constants $C',C'' > 0$, we obtain
the bound claimed in Theorem~\ref{thm:Lbound}.
We will require that $\beta \rightarrow 0$ as $n \rightarrow \infty$,
i.e., that $r = \omega(n^{-1} \log n)$.

We first establish that with high probability, the row sums of
$\Yhat$ concentrate about their expected value.
\begin{lemma} \label{lem:degconc}
Suppose that there exists constant $c_1 > 0$ such that
for all suitably large $n$ we have
\begin{equation} \label{eq:degconc:assum1}
  \frac{ \beta^2 r }{ 1 + \beta } \ge c_1 \frac{ \log n }{ n }.
\end{equation}
Then for all suitably large $n$,
with probability at least $n^{1-c_1}$,
it holds for all $i \in [n]$ that
$ |\Dhat_{ii} - \calDhat_{ii}| \le \beta \calDhat_{ii}$.
\end{lemma}
\begin{IEEEproof}
Fix $i \in [n]$. By definition,
$$
\Dhat_{ii} - \calDhat_{ii} = \sum_{j=1}^n (Y_{ij}+r) - (p\calK_{ij}+r)
	= \sum_{j=1}^n Y_{ij} - p\calK_{ij},$$
and $\E Y_{ij} = p\calK_{ij}$.
By a standard Chernoff-style bound~\cite{ChuLu2006},
$$ \Pr\left[ |\Dhat_{ii} - \calDhat_{ii}| \ge \beta \calDhat_{ii} \right]
\le 2\exp\left\{ \frac{ - 3\beta^2 \calDhat_{ii}^2 }
        { 6V + 2\beta \calDhat_{ii} } \right\}, $$
where $V = \sum_{j=1}^n \E Y_{ij}^2.$
Since
$$ V = \sum_{j=1}^n p \E K_{ij}^2
	\le p \sum_{j=1}^n \calK_{ij} \le \calDhat_{ii}, $$
we have
$$ \Pr\left[ |\Dhat_{ii} - \calDhat_{ii}| \ge \beta \calDhat_{ii} \right]
\le 2\exp\left\{ \frac{ -C \beta^2 }{1+\beta} \calDhat_{ii} \right\} , $$
where $C>0$ is a constant.
Since $\calDhat_{ii} \ge nr$ by virtue of regularization,
our assumption in~\eqref{eq:degconc:assum1} ensures that
$$ \Pr\left[ |\Dhat_{ii} - \calDhat_{ii}| \ge \beta \calDhat_{ii} \right]
	\le n^{-c_1}. $$
Applying the union bound over all $i \in [n]$ yields the result.
\end{IEEEproof}


\begin{lemma} \label{lem:hatconc:squares}
Suppose that $\gamma$ depends on $n$ in such a way that
there exist constants $C',C'' > 0$ so that for suitably large $n$,
\begin{equation} \label{eq:hatconc:squares:agrowth1}
  C' \gamma^2 \ge \frac{ 16 }{ n^2 r^3 } + \frac{ 16 }{ n^2 }
\end{equation}
and
\begin{equation} \label{eq:hatconc:squares:agrowth2}
  \gamma \ge C'' \frac{ \log^{1/2} n }{ n^{3/2} r^2 }.
\end{equation}
Then there exists a constant $c_2 > 0$ such that
with probability at least $1 - n^{-c_2}$, we have
$$ \sum_{i=1}^n \sum_{k=1}^n
        \frac{ (\Yhat_{ik}^2 - \calKhat_{ik}^2)^2 } 
        { \calDhat_{ii}^2 \calDhat_{kk}^2 } \le C \gamma^2, $$
where $C > 0 $ is a constant.
\end{lemma}
\begin{IEEEproof}
For ease of notation, define
$$ X_{ik} = \frac{ \left(\Yhat_{ik}^2 - \calKhat_{ik}^2 \right)^2 }
        { \calDhat_{ii}^2 \calDhat_{kk}^2 }. $$
We will bound
$ \Pr\left[ \sum_{i,k} X_{ik} - \E \sum_{i,k} X_{ik} \ge \gamma^2 \right] $
and show $\E \sum_{i,k} X_{ik} \le C' \gamma^2$,
implying that $ \Pr\left[ \sum_{i,k} X_{ik} \ge C \gamma^2 \right].$

A standard Chernoff-style bound lets us write
\begin{equation*}
\Pr\left[ \sum_{i,k} X_{ik} \ge \gamma^2 + \E \sum_{i,k} X_{ik} \right]
\le \exp\left\{ \frac{ -3\gamma^4 }{ 6V + 2\gamma^2 M } \right\},
\end{equation*}
where
$$ V = \sum_{i,k} \E X_{ik}^2
  = \sum_{i,k}
        \frac{ \E \left(\Yhat_{ik}^2 - \calKhat_{ik}^2 \right)^4 }
        { \calDhat_{ii}^4 \calDhat_{kk}^4 }, $$
$$
\text{and }
M = \max\left\{ 1/(\calDhat_{ii}^2 \calDhat_{kk}^2) : i,k \in [n] \right\}. $$
Bounding $V \le n^{-6}r^{-8}$ and $M \le (nr)^{-4}$,
$$
\Pr\left[ \sum_{i,k} X_{ik} \ge \gamma^2 + \E \sum_{i,k} X_{ik} \right]
\le
\exp\left\{ \frac{ -3(\gamma nr)^4 }{ 6n^{-2} r^{-4} + 2 \gamma^2 } \right\},
$$
and using our assumption in~\eqref{eq:hatconc:squares:agrowth2}
to lower bound the denominator inside the exponent by $\Omega(n\gamma^2)$,
we can guarantee the existence of a constant $c_2 > 0$ such that
$$ \Pr\left[ \sum_{i,k} X_{ik} \ge \gamma^2 + \E \sum_{i,k} X_{ik} \right]
\le
n^{-c_2}. $$

It remains for us to show that $\E \sum_{i,k} X_{ik} \le C' \gamma^2$.
We have
\begin{equation} \label{eq:hatconc:squares:main} \begin{aligned}
\E \sum_{i=1}^n & \sum_{k=1}^n X_{ik}
\le 
\sum_{i=1}^n \sum_{k=1}^n
        \frac{ \E \left( \Yhat_{ik}^4 + \calKhat_{ik}^4 \right) }
        { \calDhat_{ii}^2 \calDhat_{kk}^2 } \\
&\le
\sum_{i=1}^n \sum_{k=1}^n
        \frac{ 8\left( p\E K_{ik}^4 + r^4 \right) + \calKhat_{ik}^4 }
        { \calDhat_{ii}^2 \calDhat_{kk}^2 },
\end{aligned} \end{equation}
where we have used the fact that
$(a + b)^2 \le 2a^2 + 2b^2$ for all $a,b \in \R$.
Since $\calDhat_{ii} \ge nr$ for all $i \in [n]$,
we have
\begin{equation} \label{eq:hatconc:squares:useful}
\sum_{i=1}^n \frac{1}{\calDhat_{ii} } \le \frac{1}{ r }
\text{ and }
\sum_{i=1}^n \sum_{k=1}^n \frac{ r^4 }
        { \calDhat_{ii}^2 \calDhat_{kk}^2 } \le \frac{1}{n^2}.
\end{equation}
Noting that $\E K_{ik}^4 \le \E K_{ik} = \calK_{ik}$
and applying~\eqref{eq:hatconc:squares:useful}, we have
\begin{equation} \label{eq:hatconc:squares:bound2}
\sum_{i=1}^n \sum_{k=1}^n \frac{ p\E K_{ik}^4 }
        { \calDhat_{ii}^2 \calDhat_{kk}^2 }
\le \sum_{i=1}^n \frac{1}{\calDhat_{ii} n^2 r^2 } \le \frac{ 1 }{ n^2 r^3 }.
\end{equation}

Recalling that $\calKhat_{ik} = p\calK_{ik} + r$ by definition
and applying the definition
of $\calDhat_{ii}$,~\eqref{eq:hatconc:squares:useful} implies
\begin{equation*} \begin{aligned}
\sum_{i=1}^n & \sum_{k=1}^n \frac{ \calKhat_{ik}^4 }
        { \calDhat_{ii}^2 \calDhat_{kk}^2 }
\le 8 \sum_{i=1}^n \sum_{k=1}^n \frac{ p^4 \calK_{ik}^4 + r^4 }
        { \calDhat_{ii}^2 \calDhat_{kk}^2 } \\
&\le \frac{ 8p^3 }{ n^2 r^2 } \sum_{i=1}^n \frac{1}{ \calDhat_{ii} }
  + 8 \sum_{i=1}^n \sum_{k=1}^n \frac{ r^4 }
                        { \calDhat_{ii}^2 \calDhat_{kk}^2 } \\
&\le \frac{ 8p^3 }{n^2 r^3} + \frac{ 8 }{ n^2 }.
\end{aligned} \end{equation*}
Combining this with~\eqref{eq:hatconc:squares:main}
and~\eqref{eq:hatconc:squares:bound2}
and applying~\eqref{eq:hatconc:squares:agrowth1} completes the proof.
\end{IEEEproof}

\begin{lemma} \label{lem:hatconc:ondiag}
Under the same conditions as Lemma~\ref{lem:degconc}, 
and assuming there exists a constant $C > 0$ such that
\begin{equation} \label{eq:hatconc:ondiag:agrowth}
  C\gamma^2 \ge \frac{ \beta^2 }{ n r^2 },
\end{equation}
with probability at least $n^{1-c_1}$, we have
$$ \sum_{i=1}^n \sum_{k=1}^n \sum_{\ell=1}^n
        \frac{ (\Yhat_{ik}^2 - \calKhat_{ik}^2)
                (\Yhat_{i\ell}^2 - \calKhat_{i\ell}^2) } 
        { \calDhat_{ii}^2 \calDhat_{kk} \calDhat_{\ell\ell} }
        \le C\gamma^2. $$
\end{lemma}
\begin{IEEEproof}
Observing that $\Yhat_{ik} + \calKhat_{ik} \le 1 + p + 2r$,
\begin{equation*} \begin{aligned}
\sum_{i=1}^n & \sum_{k=1}^n \sum_{\ell=1}^n
        \frac{ (\Yhat_{ik}^2 - \calKhat_{ik}^2)
                (\Yhat_{i\ell}^2 - \calKhat_{i\ell}^2) } 
        { \calDhat_{ii}^2 \calDhat_{kk} \calDhat_{\ell\ell} } \\
&\le \frac{ (1+p + 2r)^2 }{ n^2 r^2 }
	\sum_{i=1}^n \sum_{k=1}^n \sum_{\ell=1}^n
        \frac{ (\Yhat_{ik} - \calKhat_{ik})
                (\Yhat_{i\ell} - \calKhat_{i\ell}) }
        { \calDhat_{ii}^2 }.
\end{aligned} \end{equation*}
By Lemma~\ref{lem:degconc}, with probability at least $1 - n^{1-c_1}$, 
it holds for all $i \in [n]$ that
$$
\left| \sum_{k=1}^n \Yhat_{ik} - \calKhat_{ik} \right|
	\le \beta \calDhat_{ii},
$$
and hence, since $p,r \in [0,1]$ and $\calDhat_{ii} \ge nr$,
$$ \sum_{i=1}^n \sum_{k=1}^n \sum_{\ell=1}^n
        \frac{ (\Yhat_{ik}^2 - \calKhat_{ik}^2)
                (\Yhat_{i\ell}^2 - \calKhat_{i\ell}^2) } 
        { \calDhat_{ii}^2 \calDhat_{kk} \calDhat_{\ell\ell} }
  \le \frac{ 16 \beta^2 }{ nr^2 }. $$
Our assumption in~\eqref{eq:hatconc:ondiag:agrowth} yields the desired result.
\end{IEEEproof}

\begin{lemma} \label{lem:hatconc:sumout}
$$ \sum_{i,j,k,\ell}
\frac{ p^4 \calK_{ik} \calK_{jk} \calK_{i\ell} \calK_{j\ell} }
{ \calDhat_{ii} \calDhat_{jj} \calDhat_{kk} \calDhat_{\ell\ell} }
\le \frac{p}{r}. $$
\end{lemma}
\begin{IEEEproof}
Using the following facts:
\begin{enumerate}[(i)]
\item $\calDhat_{ii} \ge rn$ for all $i \in [n]$,
\item $\calK_{ik} \in [0,1]$ for all $i,j \in [n]$,
\item $\sum_{k=1}^n p\calK_{ik} \le \calDhat_{ii}$ for all $i \in [n]$,
\end{enumerate}
we have
\begin{equation*} \begin{aligned}
\sum_{i,j,k,\ell}
\frac{ p^4 \calK_{ik} \calK_{jk} \calK_{i\ell} \calK_{j\ell} }
{ \calDhat_{ii} \calDhat_{jj} \calDhat_{kk} \calDhat_{\ell\ell} }
&\le
\frac{p}{nr} \sum_{i,j,k}
\frac{ p^2 \calK_{ik} \calK_{jk} }
{ \calDhat_{ii} \calDhat_{jj} \calDhat_{kk} }
\sum_{\ell=1}^n p \calK_{j\ell} \\
&\le \frac{p}{nr} \sum_{i,j,k}
\frac{ p^2 \calK_{ik} \calK_{jk} }{ \calDhat_{ii} \calDhat_{kk} }
\le \frac{p}{r}.
\end{aligned} \end{equation*}
\end{IEEEproof}

\begin{lemma} \label{lem:hatconc:fwvariance}
For ease of notation, let
\begin{equation} \label{eq:Xbound}
X_{ijk\ell} = \frac{ (\Yhat_{ik}\Yhat_{jk} - \calKhat_{ik}\calKhat_{jk})
        (\Yhat_{i\ell}\Yhat_{j\ell} - \calKhat_{i\ell}\calKhat_{j\ell}) } 
        { \calDhat_{ii} \calDhat_{jj} \calDhat_{kk} \calDhat_{\ell\ell} }
\end{equation}
and define
$ T = \{(i,j,k,\ell) : i,j,k,\ell \in [n] \text{ distinct.} \}$.
There exists a constant $C > 0$ such that
$$ \sum_{(i,j,k,\ell) \in T}  \VAR X_{ijk\ell} \le \frac{ C }{n^4r^5}. $$
\end{lemma}
\begin{IEEEproof}
Since $i,j,k,\ell$ are distinct for each $(i,j,k,\ell) \in T$,
\begin{equation*} \begin{aligned}
\VAR & X_{ijk\ell}
= \E X_{ijk\ell}^2 \\
&= d_{ijk\ell}^{-2}
   \E\left[ \Yhat_{ik}\Yhat_{jk} - \calKhat_{ik}\calKhat_{jk}\right]^2
   \E\left[ \Yhat_{i\ell}\Yhat_{j\ell}
		- \calKhat_{i\ell}\calKhat_{j\ell}\right]^2,
\end{aligned} \end{equation*}
where
$ d_{ijk\ell} = \calDhat_{ii} \calDhat_{jj} \calDhat_{kk} \calDhat_{\ell\ell}.$
Expanding $\Yhat_{ik} = Y_{ik} + r$ and $\calKhat_{ik} = p\calK_{ik} + r$
and using linearity of expectation, we have
\begin{equation*} \begin{aligned}
\E & \left[ \Yhat_{ik}\Yhat_{jk} - \calKhat_{ik}\calKhat_{jk}\right]^2 \\
&= \E\big[ Y_{ik}Y_{jk} - p^2\calK_{ik}\calK_{jk} \\
	&~~~~~~~~~+ r(Y_{ik} - p\calK_{ik}) + r(Y_{jk}-p\calK_{jk}) \big]^2 \\
&= \VAR Y_{ik} Y_{jk} \\
&~~~+ r(r + 2p\calK_{jk}) \VAR Y_{ik} + r(r + 2p\calK_{ik}) \VAR Y_{jk}.
\end{aligned} \end{equation*}

For ease of notation, define
$$ Q_{ijk} = p^2 \calK_{ik} \calK_{jk}
        + r(r + 2p) p\calK_{ik}
        + r(r + 2p) p\calK_{jk} . $$
The Bhatia-Davis inequality~\cite{BhaDav2000}
states that if a random variable $Z$ satisfies
$\Pr[m \le Z \le M] = 1$, then
$\VAR Z \le (\E Z - m)(M - \E Z)$.
Since $\calK_{ik} \in [0,1]$ for all $i,k \in [n]$,
we have
$ \VAR Y_{ik} Y_{jk} \le p^2 \calK_{ik} \calK_{jk}$
and $\VAR Y_{ik} \le p\calK_{ik},$ and hence
$$ \E \left[ \Yhat_{ik}\Yhat_{jk} - \calKhat_{ik}\calKhat_{jk}\right]^2
	\le Q_{ijk}. $$
Combining this with~\eqref{eq:Xbound}, we have
\begin{equation*}
\VAR X_{ijk\ell} \le d_{ijk\ell}^{-2} Q_{ijk} Q_{ij\ell}.
\end{equation*}
Summing, we have
\begin{equation*} \begin{aligned}
\sum_{(i,j,k,\ell) \in T} & \VAR X_{ijk\ell} 
\le \sum_{(i,j,k,\ell) \in T} d_{ijk\ell}^{-2} Q_{ijk} Q_{ij\ell} \\
&= \sum_{(i,j,k,\ell) \in T} d_{ijk\ell}^{-2}
	p^4 \calK_{ik} \calK_{jk} \calK_{i\ell} \calK_{j\ell}  \\
  &~~~~~~+ 4\sum_{(i,j,k,\ell) \in T} d_{ijk\ell}^{-2}
		r(r + 2p) p^3 \calK_{ik} \calK_{jk} \calK_{j\ell} \\
  &~~~~~~+ 2\sum_{(i,j,k,\ell) \in T} d_{ijk\ell}^{-2}
		r^2(r + 2p)^2 p^2 \calK_{ik} \calK_{jk} \\
  &~~~~~~+ 2\sum_{(i,j,k,\ell) \in T} d_{ijk\ell}^{-2}
	r^2(r + 2p)^2 p^2 \calK_{ik} \calK_{j\ell} \\
&\le 
\frac{ p }{ n^4 r^5 }
+ 4 \frac{ (r+2p) }{ n^4 r^4 }
+ 4 \frac{ (r + 2p)^2 }{ n^4 r^4 },
\end{aligned} \end{equation*}
where we have used
$\calDhat_{ii} \ge nr$ along with Lemma~\ref{lem:hatconc:sumout}
to bound the first sum after the equality,
and the other sums are bounded using reasoning nearly identical to
that in the proof of Lemma~\ref{lem:hatconc:sumout}.
The result then follows from $r,p \in [0,1]$.
\end{IEEEproof}

\begin{lemma} \label{lem:hatconc:fwcov}
There exists a constant $C > 0$ such that
$$ \sum_{\{(i,j,k,\ell),(i',j',k',\ell')\} \in \binom{T}{2} }
        \COV \left( X_{ijk\ell}, X_{i'k'j'\ell'} \right)
  \le  \frac{ C }{n^3r^4}. $$
\end{lemma}
\begin{IEEEproof}
Recall that
$$ X_{ijk\ell} = \frac{ (\Yhat_{ik}\Yhat_{jk} - \calKhat_{ik}\calKhat_{jk})
        (\Yhat_{i\ell}\Yhat_{j\ell} - \calKhat_{i\ell}\calKhat_{j\ell}) } 
        { \calDhat_{ii} \calDhat_{jj} \calDhat_{kk} \calDhat_{\ell\ell} } . $$
Consider first the situation
where $(a,b,c,d)$ is a permutation of $(i,j,k,\ell)$.
Call this permutation $\sigma \in S_4$.
$\sigma$ is not the identity permutation,
but $\sigma$ may be such that $X_{ijk\ell} = X_{abcd}$
as happens when, for example, $i=a,j=b,k=d,\ell=c$.
By symmetry, it suffices to consider three cases.
\paragraph{Case 1: $\{i,j\} = \{a,b\}$ }
In this case, we can assume without loss of generality (by symmetry)
that $i=b,$ $j=a$, $k=d$ and $\ell=c$, so that
\begin{equation*} \begin{aligned}
\E & X_{ijk\ell} X_{abcd}
=
\frac{
	\E\left[ (\Yhat_{ik}\Yhat_{jk} - \calKhat_{ik}\calKhat_{jk})^2
	(\Yhat_{i\ell}\Yhat_{j\ell} - \calKhat_{i\ell}\calKhat_{j\ell})^2
	\right] }
{ \calDhat_{ii}^2 \calDhat_{jj}^2 \calDhat_{kk}^2 \calDhat_{\ell\ell}^2 } \\
&= \frac{ \VAR \Yhat_{ik} \Yhat_{jk} \VAR \Yhat_{i\ell} \Yhat_{j\ell} }
{ \calDhat_{ii}^2 \calDhat_{jj}^2 \calDhat_{kk}^2 \calDhat_{\ell\ell}^2 }
\le \frac{ (1+r)^4
	\calKhat_{ik} \calKhat_{jk} \calKhat_{i\ell} \calKhat_{j\ell} }
{ \calDhat_{ii}^2 \calDhat_{jj}^2 \calDhat_{kk}^2 \calDhat_{\ell\ell}^2 },
\end{aligned} \end{equation*}
where the last inequality follows from the Bhatia-Davis inequality
and the fact that $0 \le \Yhat_{ik} \le 1+r$.

\paragraph{Case 2: $\{i,j\} = \{a,c\}$ }
Without loss of generality, assume that
$i=a$, $j=c$, $k=b$ and $\ell = d$. We have
\begin{equation*} \begin{aligned}
\E X_{ijk\ell} X_{abcd}
&=
\frac{ \calKhat_{ik} \calKhat_{ij} \calKhat_{j\ell} \calKhat_{k\ell}
	\VAR \Yhat_{jk} \VAR \Yhat_{i\ell} }
{ \calDhat_{ii}^2 \calDhat_{jj}^2 \calDhat_{kk}^2 \calDhat_{\ell\ell}^2 } \\
&\le 
\frac{ (1+r)^2 \calKhat_{ik} \calKhat_{j\ell} \calKhat_{jk} \calKhat_{i\ell} }
{ \calDhat_{ii}^2 \calDhat_{jj}^2 \calDhat_{kk}^2 \calDhat_{\ell\ell}^2 },
\end{aligned} \end{equation*}
where the inequality follows from the Bhatia-Davis inequality and
the fact that $\calKhat_{ik} \le 1+r$.

\paragraph{Case 3: $\{i,j\} = \{c,d\}$ }
Without loss of generality, assume that
$i=c$, $j=d$, $k=a$ and $\ell = b$. Then
\begin{align*}
\E & X_{ijk\ell} X_{abcd} \\
&=
\frac{ 
\E \Yhat_{ik} \Yhat_{j\ell} (\Yhat_{jk} + \Yhat_{i\ell})^2
- \calKhat_{ik} \calKhat_{j\ell}(\calKhat_{jk} + \calKhat_{i\ell})^2 }
{ \calDhat_{ii}^2 \calDhat_{jj}^2 \calDhat_{kk}^2 \calDhat_{\ell\ell}^2 } \\
&=
\frac{ \calKhat_{ik}\calKhat_{j\ell}
	\left( \VAR \Yhat_{jk} + \VAR \Yhat_{i\ell} \right) }
{ \calDhat_{ii}^2 \calDhat_{jj}^2 \calDhat_{kk}^2 \calDhat_{\ell\ell}^2 }.
\end{align*}
					
Letting $(i,j,k,\ell) \sim (a,b,c,d)$ denote the fact that
$(a,b,c,d)$ is a permutation of $(i,j,k,\ell)$, we can bound 
the sum of the covariances under consideration by
\begin{equation} \label{eq:permubound} \begin{aligned}
\sum_{(i,j,k,\ell) \in T}
&\sum_{(a,b,c,d) \sim (i,j,k,\ell)}
\COV\left( X_{ijk\ell}, X_{abcd} \right) \\
&\le 2C(1+r)^4 \sum_{i,j,k,\ell}
	\frac{ \calKhat_{ik} \calKhat_{jk} \calKhat_{i\ell} \calKhat_{j\ell} }
{ \calDhat_{ii}^2 \calDhat_{jj}^2 \calDhat_{kk}^2 \calDhat_{\ell\ell}^2 } \\
&~~~~~~+ 2C(1+r) \sum_{(i,j,k,\ell) \in T}
	\frac{ \calKhat_{ik}\calKhat_{j\ell}\calKhat_{jk} }
{ \calDhat_{ii}^2 \calDhat_{jj}^2 \calDhat_{kk}^2 \calDhat_{\ell\ell}^2 } \\
&\le \frac{ C(1+r)^5 + 2C(1+r) }{ n^4 r^5 },
\end{aligned} \end{equation}

Now, consider the situation where $(i,j,k,\ell)$ is not a permutation
of $(a,b,c,d)$.
Clearly, if $\{i,j,k,\ell\} \cap \{a,b,c,d\} = \emptyset$,
then $\COV(X_{ijk\ell}, X_{abcd}) = 0$.
Indeed, $\COV(X_{ijk\ell},X_{abcd}) \neq 0$ requires that
each term of the form $(\Yhat_{ik}\Yhat_{jk} - \calKhat_{ik}\calKhat_{jk})$
be dependent on one of the other three such terms in $X_{ijkl}X_{abcd}$,
since otherwise a term of the form
$\E(\Yhat_{ik}\Yhat_{jk} - \calKhat_{ik}\calKhat_{jk})$ factors out
and the covariance is zero.
Indeed, only one other choice (up to permutations of the indices)
of $(i,j,k,\ell)$ and $(a,b,c,d)$ gives rise to a non-zero covariance,
namely $\E X_{ijk\ell} X_{ibk\ell}$.
By symmetry, to handle the terms of this form, it will suffice for us to bound
$$ \sum_{(i,j,k,\ell) \in T} \sum_{b \not \in \{i,j,k,\ell\} }
	\COV(X_{ijk\ell}, X_{ibk\ell}). $$
Using the fact that $\VAR \Yhat_{ik} \le \calKhat_{ik}$
by the Bhatia-Davis inequality,
and applying reasoning similar to that in Lemma~\ref{lem:hatconc:sumout},
\begin{equation*} \begin{aligned}
&\sum_{(i,j,k,\ell) \in T} \sum_{b \not \in \{i,j,k,\ell\} }
	\COV(X_{ijk\ell}, X_{ibk\ell}) \\
&= \sum_{(i,j,k,\ell) \in T} \sum_{b \not \in \{i,j,k,\ell\} }
	\frac{ \calKhat_{jk} \calKhat_{bk}
		\calKhat_{j\ell} \calKhat_{b\ell}
		\VAR \Yhat_{ik} \VAR \Yhat_{i \ell} }
	{ \calDhat_{ii}^2 \calDhat_{kk}^2 \calDhat_{\ell\ell}^2
		\calDhat_{jj} \calDhat_{bb} } \\
&\le \sum_{(i,j,k,\ell) \in T} \sum_{b \not \in \{i,j,k,\ell\} }
        \frac{ \calKhat_{jk} \calKhat_{bk} \calKhat_{j\ell} \calKhat_{b\ell}
		\calKhat_{ik} \calKhat_{i \ell} }
	{ \calDhat_{ii}^2 \calDhat_{kk}^2 \calDhat_{\ell\ell}^2
		\calDhat_{jj} \calDhat_{bb} } \\
&\le \frac{(1+r)^2}{(nr)^4}
	\sum_{i,j,k \in [n] \text{ distinct}}
        \frac{ \calKhat_{jk} \calKhat_{ik} }
	{ \calDhat_{kk}^2 } \le \frac{(1+r)^2}{n^3r^4}.
\end{aligned} \end{equation*}

Combining this with~\eqref{eq:permubound} and
noting that $r > n^{-1}$ implies $(n^3 r^4)^{-1} \ge (n^4 r^5)^{-1}$,
we have our result.
\end{IEEEproof}

\begin{lemma} \label{lem:hatconc:fourwise}
Let $ T = \{(i,j,k,\ell) : i,j,k,\ell \in [n] \text{ distinct.} \}$.
For each $(i,j,k,\ell) \in T$, define variable
$$ X_{ijk\ell} = 
\frac{ (\Yhat_{ik}\Yhat_{jk} - \calKhat_{ik}\calKhat_{jk})
        (\Yhat_{i\ell}\Yhat_{j\ell} - \calKhat_{i\ell}\calKhat_{j\ell}) } 
        { \calDhat_{ii} \calDhat_{jj} \calDhat_{kk} \calDhat_{\ell\ell} }. $$
There exist constants $C, C_\gamma > 0$ such that
with probability at least $1 - C_\gamma (\gamma^4n^3r^4)^{-1}$,
\begin{equation}
\sum_{(i,j,k,\ell) \in T}
\frac{ (\Yhat_{ik}\Yhat_{jk} - \calKhat_{ik}\calKhat_{jk})
        (\Yhat_{i\ell}\Yhat_{j\ell} - \calKhat_{i\ell}\calKhat_{j\ell}) } 
        { \calDhat_{ii} \calDhat_{jj} \calDhat_{kk} \calDhat_{\ell\ell} }
\le C \gamma^2.
\end{equation}
\end{lemma}
\begin{IEEEproof}
By Chebyshev's inequality,
$$ \Pr\left[ \sum_{(i,j,k,\ell) \in T} X_{ijk\ell} \ge C \gamma^2 \right]
   \le \frac{ \VAR \sum_{(i,j,k,\ell) \in T} X_{ijk\ell} }{ C^2 \gamma^4 }. $$
We have
\begin{equation*} \begin{aligned}
\VAR & \sum_{(i,j,k,\ell) \in T} X_{ijk\ell} \\
&= \sum_{(i,j,k,\ell) \in T} \VAR X_{ijk\ell} \\
&~~~+ \sum_{\{(i,j,k,\ell),(i',j',k',\ell')\} \in \binom{T}{2} }
	\COV \left( X_{ijk\ell}, X_{i'k'j'\ell'} \right).
\end{aligned} \end{equation*}
Lemma~\ref{lem:hatconc:fwvariance} bounds the first of these two sums by
$$ \sum_{(i,j,k,\ell) \in T} \VAR X_{ijk\ell}
  \le \frac{ C' }{n^4r^5}, $$
where $C' > 0$ is a constant,
and Lemma~\ref{lem:hatconc:fwcov} ensures that
$$ \sum_{\{(i,j,k,\ell),(i',j',k',\ell')\} \in \binom{T}{2} }
	\COV \left( X_{ijk\ell}, X_{i'k'j'\ell'} \right)
  \le \frac{ C'' }{n^3r^4} $$
for some constant $C'' > 0$.
Since $(n^4r^5)^{-1} \le (n^3r^4)^{-1}$ for $r > 1/n$, we have
$$ \Pr\left[ \sum_{(i,j,k,\ell) \in T} X_{ijk\ell} \ge C \gamma^2 \right]
	\le \frac{ C' + C'' }{ C \gamma^4 n^3 r^4 }. $$
Choosing $C_\gamma = (C'+C'')/C$ yields the result.
\end{IEEEproof}

\begin{lemma} \label{lem:hatclose}
Under the conditions of the above lemmata,
there exist constants $c,C > 0$ such that
for all suitably large $n$,
with probability at least $1-3n^{-c}$, we have
$$ \| \calLhat\calLhat
                - (\calDhat^{-1/2}\Yhat \calDhat^{-1/2})^2 \|_F \le C\gamma. $$
\end{lemma}
\begin{IEEEproof}
Expanding the sum and recalling our earlier definition of
$T = \{(i,j,k,\ell): i,j,k,\ell \in [n] \text{ distinct.} \}$, we have
\begin{equation*} \begin{aligned}
\| & \calLhat\calLhat
                - (\calDhat^{-1/2}\Yhat \calDhat^{-1/2})^2 \|_F^2 \\
&= \sum_{i,j,k,\ell}
   \frac{ (\Yhat_{ik}\Yhat_{jk} - \calKhat_{ik}\calKhat_{jk})
	  (\Yhat_{i\ell}\Yhat_{j\ell} - \calKhat_{i\ell}\calKhat_{j\ell} ) }
	{ \calDhat_{ii} \calDhat_{jj} \calDhat_{kk} \calDhat_{\ell\ell} } \\
&=
\sum_{i=1}^n \sum_{k \neq i}
	\frac{(\Yhat_{ik}^2 -\calKhat_{ik}^2)^2 }
	{ \calDhat_{ii}^2 \calDhat_{kk}^2 } \\
&~~~+ 
\sum_{i=1}^n \sum_{k \neq i} \sum_{\ell \neq i }
	\frac{(\Yhat_{ik}^2 -\calKhat_{ik}^2)
		( \Yhat_{i\ell}^2 -\calKhat_{i\ell}^2  ) }
        { \calDhat_{ii}^2 \calDhat_{kk} \calDhat_{\ell\ell} } \\
&~~~+ 
\sum_{(i,j,k,\ell) \in T}
	\frac{(\Yhat_{ik}\Yhat_{jk} -\calKhat_{ik}\calKhat_{jk})
	( \Yhat_{i\ell}\Yhat_{j\ell} -\calKhat_{i\ell}\calKhat_{j\ell} ) }
        { \calDhat_{ii} \calDhat_{jj} \calDhat_{kk} \calDhat_{\ell\ell} } .	
\end{aligned} \end{equation*}
Each of these three summations is bounded 
(with high probability) by $C\gamma^2$
by Lemmata~\ref{lem:hatconc:squares},
\ref{lem:hatconc:ondiag} and \ref{lem:hatconc:fourwise}, respectively.
Let constants $c_1,c_2 > 0$ be as defined in
Lemma~\ref{lem:degconc} and
Lemma~\ref{lem:hatconc:squares} respectively,
and choose $c_3 > 0$ so that
$C_\gamma( \gamma^4n^3r^4)^{-1} \le n^{-c_3}$
for suitably large $n$,
where $C_\gamma$ is as defined in
Lemma~\ref{lem:hatconc:fourwise}.
By the union bound, with probability at least
$1 - (n^{1-c_1} + n^{-c_2} + n^{-c_3})$,
all three sums are bounded at once, and the result follows
by taking $c = \min\{ c_1-1, c_2, c_3 \}$,
\end{IEEEproof}

\begin{lemma} \label{lem:calclose}
Suppose that $\beta \rightarrow 0$ as $n \rightarrow \infty$.
Under the conditions of Lemma~\ref{lem:degconc},
there exists a constant $C > 0$ such that
with probability at least $1 - n^{1-c_1}$,
$$ \| \Lhat\Lhat
                - (\calDhat^{-1/2}\Yhat \calDhat^{-1/2})^2 \|_F
  \le C\frac{ \beta }{ r^{1/2} }. $$
\end{lemma}
\begin{IEEEproof}
Under the conditions of Lemma~\ref{lem:degconc},
with probability at least $1-n^{1-c_1}$ it holds
for all $i \in [n]$ that
$|\calDhat_{ii} - \sum_{k=1}^n \Yhat_{ik}| \le \beta \calDhat_{ii}$.
It follows
that for a suitably chosen constant $C'>0$,
for all $i,j,k \in [n]$ we have
\begin{equation} \label{eq:RCYbetabound} 
\left| \frac{1}{ \Dhat^{1/2}_{ii} \Dhat^{1/2}_{jj} \Dhat_{kk} }
                - \frac{1}
                {\calDhat^{1/2}_{ii} \calDhat^{1/2}_{jj} \calDhat_{kk} }
        \right|
\le \frac{ C'\beta }
	{ \calDhat^{1/2}_{ii} \calDhat^{1/2}_{jj} \calDhat_{kk} }.
\end{equation}
To see why this is the case
(here we are following the argument motivating Equation
A.6 in~\cite{RohChaYu2011}),
note that when
$|\calDhat_{ii} - \sum_{k=1}^n \Yhat_{ik}| \le \beta \calDhat_{ii}$
for all $i \in [n]$, we have
$$ 
\frac{(1+\beta)^{-2}}{\calDhat^{1/2}_{ii} \calDhat^{1/2}_{jj} \calDhat_{kk} }
\le
\frac{1}{ \Dhat^{1/2}_{ii} \Dhat^{1/2}_{jj} \Dhat_{kk} }
\le
\frac{(1-\beta)^{-2}}{\calDhat^{1/2}_{ii} \calDhat^{1/2}_{jj} \calDhat_{kk} },
$$
and Equation~\ref{eq:RCYbetabound} follows, since
$\beta \rightarrow 0$ as $n \rightarrow \infty$, and thus
\begin{equation*} \begin{aligned}
(1+\beta)^{-2} &\ge \frac{\beta^{-2} - 1}{(\beta^{-1}+1)^2}
= \frac{\beta^{-1} - 1}{\beta^{-1} + 1}
\ge 1 - C''\beta, \\
(1-\beta)^{-2} &= 1 + \frac{2}{ \beta^{-1} - 1} + \frac{1}{(\beta^{-1} - 1)^2 }
\le 1 + C''\beta.
\end{aligned} \end{equation*}

Using~\eqref{eq:RCYbetabound}, we have
\begin{equation*} 
\| \Lhat\Lhat - (\calDhat^{-1/2}\Yhat \calDhat^{-1/2})^2 \|_F^2
\le C'\beta^2 \sum_{i,j,k,\ell}
	\frac{ \Yhat_{ik} \Yhat_{jk} \Yhat_{i\ell} \Yhat_{j\ell} }
	{ \calDhat_{ii} \calDhat_{jj} \calDhat_{kk} \calDhat_{\ell\ell} }.
\end{equation*}
Under the same event, we have
$\sum_{k=1}^n \Yhat_{ik} \le (1+\beta) \calDhat_{ii}$ for all $i\in[n]$,
and making repeated use of this and the facts that $\Yhat_{jk} \le (1+r)$,
and $\calDhat_{ii} \ge nr$, it follows that
\begin{equation*} \begin{aligned}
\| \Lhat\Lhat &- (\calDhat^{-1/2}\Yhat \calDhat^{-1/2})^2 \|_F^2
\le C' \beta^2 \sum_{i,j,k,\ell} 
        \frac{ \Yhat_{ik} \Yhat_{jk} \Yhat_{i\ell} \Yhat_{j\ell} }
        { \calDhat_{ii} \calDhat_{jj} \calDhat_{kk} \calDhat_{\ell\ell} } \\
&\le \frac{ \beta^2(1+r)(1+\beta)^3 }{ r }.
\end{aligned} \end{equation*}
The result follows since $r$ and $\beta$ are bounded above by $1$.
\end{IEEEproof}

To obtain our result in Theorem~\ref{thm:Lbound},
take $\gamma = C' n^{-1/2} r^{-1} \log^{1/2} n$
and $\beta = C'' n^{-1/2} r^{-1/2} \log^{1/2} n$
for suitably large constants $C',C'' > 0$.
Note first that these choices of $\gamma$ and $\beta$
satisfy all of the constraints of the lemmata
required for Lemma~\ref{lem:hatclose},
so long as $r = \omega( n^{-1} \log n)$.
Further, note that $\beta/r^{1/2} = C \gamma$
for some constant $C > 0$, and hence
Lemma~\ref{lem:calclose} implies that
$\| \Lhat\Lhat - (\calDhat^{-1/2}\Yhat \calDhat^{-1/2})^2 \|_F \le C\gamma$
with high probability.
Combining Lemma~\ref{lem:hatclose} and Lemma~\ref{lem:calclose}
and applying the triangle inequality then yields Theorem~\ref{thm:Lbound}.

\bibliographystyle{plain}
\bibliography{biblio}

\bibliographystyle{IEEEtran}
\bibliography{biblio}

%

\begin{IEEEbiography}{Keith Levin}
Biography text here.
\end{IEEEbiography}

\begin{IEEEbiographynophoto}{Vince Lyzinski}
Biography text here.
\end{IEEEbiographynophoto}

\end{document}